\documentclass[11pt]{article}

\usepackage[final]{acl}

\usepackage{times}
\usepackage{latexsym}

\usepackage[T1]{fontenc}

\usepackage[utf8]{inputenc}

\usepackage{microtype}

\usepackage{inconsolata}

\usepackage{graphicx}

\usepackage{booktabs}
\usepackage{multirow}
\usepackage{CJK}
\usepackage{makecell}
\usepackage{pifont}
\usepackage{array}

\usepackage{amsmath}
\usepackage{booktabs}

\usepackage{rotating}
\usepackage{bm}
\usepackage{arydshln}

\title{RECAP: Resistance Capture in Text-based Mental Health Counseling with Large Language Models}

\author{
 \textbf{Anqi Li\textsuperscript{1,2}},
 \textbf{Yuqian Chen\textsuperscript{2}},
 \textbf{Yu Lu\textsuperscript{2}},
 \textbf{Zhaoming Chen\textsuperscript{2}},
\\
 \textbf{Yuan Xie\textsuperscript{2}\textsuperscript{*}},
 \textbf{Zhenzhong Lan\textsuperscript{2}\textsuperscript{*}},
\\
 \textsuperscript{1}Zhejiang University,
 \textsuperscript{2}Westlake University \\
 \textsuperscript{*}Corresponding authors
\\
 \small{
   \href{}{lianqi, xieyuan, lanzhenzhong@westlake.edu.cn}
 }
}

\begin{document}
\begin{CJK*}{UTF8}{gbsn}

\maketitle
\begin{abstract}
Recognizing and navigating client resistance is critical for effective mental health counseling, yet detecting such behaviors is particularly challenging in text-based interactions. Existing NLP approaches oversimplify resistance categories, ignore the sequential dynamics of therapeutic interventions, and offer limited interpretability.

To address these limitations, we propose~\textbf{PsyFIRE}, a theoretically grounded framework capturing 13 fine-grained resistance behaviors alongside collaborative interactions. Based on PsyFIRE, we construct the~\textbf{ClientResistance} corpus with 23,930 annotated utterances from real-world Chinese text-based counseling, each supported by context-specific rationales. Leveraging this dataset, we develop~\textbf{RECAP}, a two-stage framework that detects resistance and fine-grained resistance types with explanations. 

RECAP achieves 91.25\% F1 for distinguishing collaboration and resistance and 66.58\% macro-F1 for fine-grained resistance categories classification, outperforming leading prompt-based LLM baselines by over 20 points. Applied to a separate counseling dataset and a pilot study with 62 counselors, RECAP reveals the prevalence of resistance, its negative impact on therapeutic relationships and demonstrates its potential to improve counselors' understanding and intervention strategies.

\end{abstract}

\section{Introduction}

\begin{figure}
    \centering
    \includegraphics[width=1\linewidth]{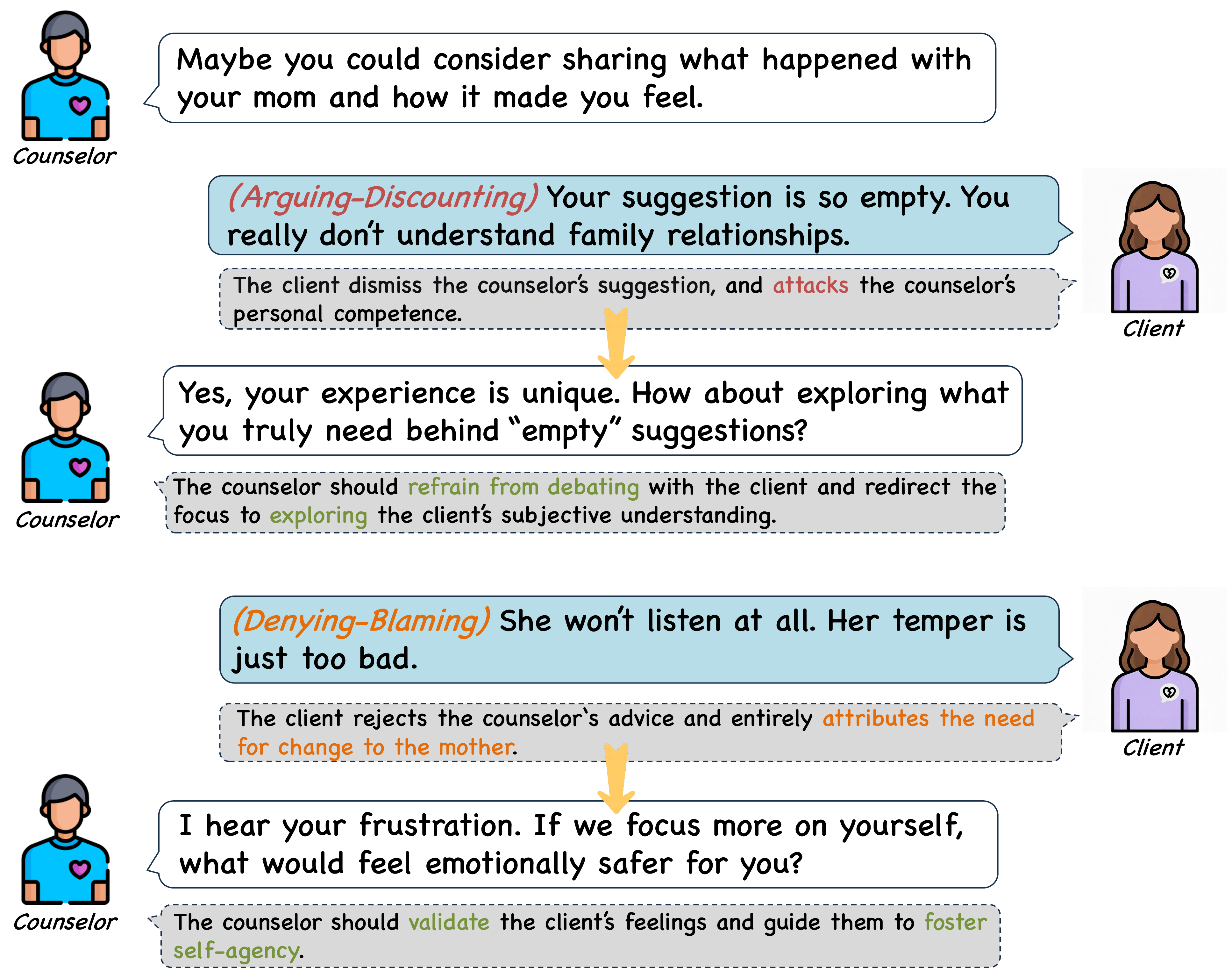}
    \caption{A counseling dyad example showing that identical counselor interventions can elicit different client resistance behaviors, each requiring tailored counselor responses.}
    \label{fig:placeholder}
\end{figure}


Timely recognition of client resistance to counselors' interventions is clinically crucial, as it enables more targeted responses and helps prevent stalled progress or premature dropout~\citep{hara2015therapist, aviram2016responsive, greenberg1986change, leahy2008therapeutic}. However, detecting resistance in text-based counseling is particularly challenging for counselors, where clients often express it subtly and in diverse ways through written language alone~\citep{hill2001relation, watson2000alliance}. For instance, clients may deflect from emotionally difficult topics with vague remarks or respond to suggestions with hopelessness or resignation~\citep{beutler2001resistance, urmanche2019ambivalence}. If these signals are not identified and addressed in a timely manner, they can disrupt the therapeutic process and undermine treatment effectiveness~\citep{CHAMBERLAIN1984144CRC, beutler2002resistance}.

To assist counselors in identifying and interpreting client resistance, prior work has applied natural language processing (NLP) methods to to automatically recognize resistance in counseling dialogues~\citep{tanana-etal-2015-recursive, cao2019observing, li-etal-2023-understanding}. However, existing approaches suffer from three key limitations: (1)~\textbf{Oversimplified classification}. Diverse resistance behaviors are aggregated into broad categories such as self-defensiveness~\citep{tanana-etal-2015-recursive, cao2019observing, li-etal-2023-understanding}, obscuring meaningful distinctions and limiting scenario-specific guidance; (2)~\textbf{Intervention-agnostic modeling}. Most methods categorize client utterances based on superficial linguistic patterns, such as \textit{verbosity}, without modeling the sequential dynamics through which resistance emerges in response to specific therapeutic interventions~\citep{otani1989client, park-etal-2019-conversation}; and (3)~\textbf{Lack of explainable outputs.} Predictions are typically provided without supporting rationales, reducing interpretability and practical utility.


In this paper, we propose a novel computational approach to capture clients' diverse resistance behaviors as responses to counselors' therapeutic interventions, together with supporting rationales. We first introduce \textbf{PsyFIRE}\footnote{\textbf{FI}ne-grained \textbf{RE}sistance in text-based \textbf{Psy}chological Counseling}, a theoretically grounded annotation framework developed in collaboration with psychological experts. PsyFIRE adapts established resistance constructs from face-to-face therapy~\citep{chamberlain1984observation,ribeiro2014TCCS} to text-based counseling and defines 13 fine-grained resistance behaviors, along with a collaborative category (\S\ref{section: framework}). Building on PsyFIRE, we construct \textbf{ClientResistance}, a large-scale corpus of 23,930 client utterances from real-world Chinese text-based counseling sessions, annotated with fine-grained resistance labels and context-specific rationales (\S\ref{sec: data}). Using this corpus, we develop \textbf{RECAP}, a two-stage, explainable framework for detecting client resistance and fine-grained resistance behaviors. We fine-tune Llama-3.1-8B-Instruct on annotated samples with explanations (\S\ref{sec: model}) using 5-fold cross validation. RECAP effectively distinguishes resistance from collaboration, achieving an F1 score of 91.25\% and demonstrating high sensitivity to resistance with nearly 90\% recall. It also accurately captures fine-grained resistance behaviors, reaching 67.72\% accuracy and 66.58\% macro-F1, surpassing leading prompt-based LLM baselines like GPT-4o and Claude-3.5-Sonnet by over 20 points (\S\ref{sec: results}).


Applying RECAP to a separate real-world counseling dataset~\citep{li2024automaticevaluationmentalhealth}, we find that client resistance is both pervasive and multifaceted. Higher resistance levels are strongly associated with weaker therapeutic relationships, with specific behaviors (e.g., \textit{Discounting}) showing particularly detrimental effects. These findings reveal key opportunities for delivering targeted feedback to counselors. Moreover, a proof-of-concept experiment with 62 counselors demonstrates that real-time identification of resistance types, along with RECAP-generated explanations, substantially improves counselors' understanding of client behaviors and supports more effective intervention strategies (\S\ref{sec: application}).

\textbf{Our contributions can be summarized as:}

\noindent 1. We introduce PsyFIRE, a theoretically grounded framework that defines 13 fine-grained client resistance behaviors in text-based counseling

\noindent 2. We construct ClientResistance, a large-scale annotated dataset of 23,930 real-world utterances with expert rationales.

\noindent 3. We propose RECAP, an explainable framework for resistance detection and fine-grained classification, achieving substantial gains over strong LLM baselines.

\noindent 4. We validate the practical utility of RECAP through analyses on independent counseling data and a proof-of-concept study with counselors, showing its value for understanding and responding to client resistance.

\section{Related Work}

\begin{table*}[]
\centering
\scalebox{0.7}{

\begin{tabular}{cl|c|c|c|c|c}
\toprule

\multicolumn{2}{c}{\multirow{3}{*}{}}                                       & \multirow{3}{*}{\textbf{Context}}   & \multirow{3}{*}{\textbf{\makecell[c]{Applicable to \\ Text-based \\ Counseling}}} & \multicolumn{3}{c}{\textbf{Measurement Dimension}}    \\
\multicolumn{2}{c}{}    &    &    & \multirow{2}{*}{\textbf{Interactional}} & \multirow{2}{*}{\textbf{Moment-by-Moment}} & \multirow{2}{*}{\textbf{Fine-grained}} \\
\multicolumn{2}{c}{}       &    &  & &    &       \\

\midrule
\multirow{5}{*}{\rotatebox[origin=c]{90}{Scales}}  & ~\citep{schuller1991resistance}   & Voice-based therapy  & \ding{55}  & \ding{55}  & \ding{55}   &  \ding{51}            \\
& ~\citep{mahalik1994development}   &  Face-to-face therapy  & \ding{55} & \ding{55} & \ding{55} & \ding{51} \\
& ~\citep{otani1989client}  &  Face-to-face therapy    & \ding{55}     & \ding{55}    & \ding{51}               & \ding{51} \\
& ~\citep{chamberlain1984observation}    &    Face-to-face therapy   & \ding{55}   & \ding{51}         & \ding{51}  & \ding{51} \\
& ~\citep{ribeiro2014TCCS}    & Face-to-face therapy & \ding{55}                                  & \ding{51}         & \ding{51}               & \ding{51}           \\
\midrule
\multirow{4}{*}{\rotatebox[origin=c]{90}{Methods}} & ~\citep{shoham1989paradoxicalinterventions} & Voice-based therapy  & \ding{55}   &     \ding{55}       & \ding{55}               &    \ding{55}   \\
& ~\citep{sarma2015towards} & Motivational Interview     &   \ding{55}   &    \ding{51}        & \ding{51}               & \ding{55} \\
& ~\citep{lv2021psychological}                & Voice-based therapy  & \ding{55}                                  & \ding{55}         & \ding{51}               & \ding{51}           \\
& ~\citep{li-etal-2023-understanding}                                 & Text-based counseling      & \ding{51}                                  & \ding{51}         & \ding{51}               & \ding{55}   \\ 
\midrule
& RECAP & Text-based counseling & \ding{51} & \ding{51} & \ding{51} & \ding{51} \\
\bottomrule
\end{tabular}}
\caption{Comparison with existing approaches. Unlike prior work limited to face-to-face therapy, RECAP captures interactive and fine-grained client resistance behaviors on a moment-by-moment basis in text-based counseling.}
\label{tab:comparison_related_work}
\end{table*}

\paragraph{How to Measure Resistance?}



Table~\ref{tab:comparison_related_work} summarizes existing psychotherapy measurement tools to capture client resistance during counseling as an in-session state. Scales introduced by~\citet{schuller1991resistance} and~\citet{mahalik1994development} assess clients' overall resistance level throughout the counseling process, while those proposed by~\citet{chamberlain1984observation} and~\citet{ribeiro2014TCCS} are designed to track moment-by-moment resistance behaviors within sessions. However, these tools were originally designed for face-to-face counseling and rely heavily on non-verbal cues such as facial expressions and vocal tone. In contrast, in text-based counseling, resistance must be inferred solely from written responses. Moreover, some scales treat client resistance as a static, unilateral behavior, overlooking its interpersonal and dynamically co-constructed nature~\citep{otani1989client}. As a result, labels such as "verbosity" provide limited insight into clients' underlying motivations, reducing their practical value for guiding therapeutic responses.

To address these limitations, we adapted existing resistance frameworks to the text-based counseling context, emphasizing moment-to-moment and interactional measurement. We introduce a refined taxonomy tailored for text-based interactions, compile a newly annotated dataset of resistance behaviors, propose a computational approach for resistance identification, and discuss its practical implications for mental health platforms.


\paragraph{Computational Approaches for Resistance Detection.} 


To our knowledge, most recent computational approaches for detecting client resistance in counseling dialogues have been designed primarily for spoken contexts, relying on acoustic features such as vocal pitch cues. These methods are therefore unsuitable for application in online text‑based counseling settings~\citep{lv2021psychological, shoham1989paradoxicalinterventions}. Among the few NLP studies targeting resistance detection in written conversations, existing classifications remain coarse-grained, typically limited to binary distinctions between cooperative and non‑cooperative responses~\citep{sarma2015towards, li-etal-2023-understanding}. As different forms of resistance often reflect distinct underlying client motivations, such a simplified scheme constrains therapists' ability to design tailored therapeutic interventions. In addition, most existing approaches fail to provide rationale or interpretable explanations behind their predictions, limiting the practical utility in offering meaningful guidance to therapists.

To address these gaps, this study proposes a computational framework that identifies varied client resistance in text-based psychotherapy and generates interpretable explanations.

\section{Resistance Taxonomy}
\label{section: framework}
To better understand client resistance in text-based counseling conversations, we introduce PsyFIRE, an innovative framework to categorize clients' non-collaborative behaviors towards therapists. More details are shown in Appendix~\ref{appendix: framework_development}.

\subsection{Development Process}

Through collaboration with experts in counseling psychology, we have integrated and adapted existing taxonomies designed for face-to-face counseling, including the Client Resistance Observation Scheme~\citep{CHAMBERLAIN1984144CRC}, its adaptation for motivational interviewing~\citep{tip35_mi_1999}, and the Client Resistance in Counseling~\citep{otani1989client}, to the context of online text-based psychological counseling.

Three developers\footnote{One is a postdoctoral researcher specializing in psychology and conversation analysis; another is a postgraduate in clinical psychology with hospital counseling experience; and the third is a doctoral student working at the intersection of computer science and clinical psychology.} constructed the framework using a consensual qualitative research approach~\citep{hill1997CQR}. Each client utterance was assigned a single, primary resistance category to enhance inter-rater consistency and offer counselors clear, actionable guidance for intervention~\citep{carletta-etal-1997-reliability, sajid2023single}.

\subsection{Taxonomy}

\begin{table*}[]
\centering
\renewcommand{\arraystretch}{1.2} 
\scalebox{0.69}{
\begin{tabular}{m{2cm}<{\centering}|m{2.5cm}<{\centering}|m{9cm}<{\centering}|m{7.8cm}<{\centering}}
\toprule
\multicolumn{2}{c}{\textbf{Behaviors}}     & \textbf{Examples}                                       & \textbf{Lexical Features}  \\
\hline
\multirow{3}{*}{Arguing}   & Challenging   & \textit{Do you really believe that staying calm is even possible when someone is constantly provoking you?}  &  counseling (11.02), you (8.77), suggestion (6.71), counselor (6.40), chat (6.00)  \\ \cline{2-4}

& Discounting   & \textit{Is "just communicate" the only trick you therapists have?}  &  confusion (5.32), freedom (5.13), conversation (4.47), relationship (4.09), develop (4.07)    \\ 
\hline
\multirow{12}{*}{Denying}   & Blaming       & \textit{It's my husband who's always so emotional; he's the one who really needs to work on his communication skills.}       & communicate (9.39), useless (8.19), angry (5.58), parents (4.46), control (4.38)     \\ \cline{2-4}

& Disagreeing   & \textit{I think this approach might not work for me.} & but (5.82), accept (4.66), relax (4.61), care (4.34), regret (3.96)  \\ \cline{2-4}

& Excusing      & \textit{I have been quite busy with work recently and may not have time to have a proper conversation with him.}   &    mobile phone (5.76), space (5.14), sports (5.14), awkward (4.83), work (4.62) \\ \cline{2-4}

& Minimizing    & \textit{I don't think the issues between us are serious enough to warrant such a formal resolution.} & not bad (9.58), nothing much (5.62), it's fine (5.53), be used to (5.48), hard to say (4.03) \\ \cline{2-4}

& Pessimism     & \textit{I'm probably just someone with terrible emotional control. My life isn't going to get any better anyway.}   & ability (5.24), not good enough (5.13), self (4.97), seems like (4.53), strength (4.15)  \\ \cline{2-4}

& Reluctance    & \textit{I understand what I should do, but I need to think it over.} & perhaps (8.79), think it over (7.49), so difficult (6.38), think (5.06), a little (4.85) \\ \cline{2-4}

& Unwillingness & \textit{I just don't want to try that method.} & don't wanna (16.18), change (8.37), communicate (5.30), forget it (4.87), troublesome (4.56) \\ \cline{2-4}
\hline

\multirow{3}{*}{Avoidance} & Minimum Talk  & \textit{Okay...}  & nothing (10.89), nothing special (6.84), special (6.37), hard to explain (5.44), just okay (5.27) \\ \cline{2-4}

& Limit Setting & \textit{Can we not talk about this today? Let's discuss something else.}  &  don't want to (11.23), anyway (7.33), don't really want to (6.11), answer (5.47), for now (5.45)                  \\
\hline

\multirow{3}{*}{Ignoring}  & Sidetracking & \textit{Speaking of communication, my child has been having issues at school recently...} & teacher (7.91), sister (5.89), yesterday (5.62), girls (5.11), time (4.63) \\ \cline{2-4}

& Inattention   & \textit{He also said some very hurtful things...} (\textit{continues venting about arguments with husband}) &   classmates (5.12), children (4.34), take care (4.26), families (4.07), hospital (4.04)     \\  

\bottomrule                                             
\end{tabular}}
\caption{Overview of the proposed PsyFIRE taxonomy. The framework comprises four coarse-grained behavioral patterns and thirteen fine-grained categories of client resistance in text-based counseling. The "Lexical Features" column presents the top five associated unigrams and bigrams for each category, ranked by their rounded z-scored log odds ratios \citep{monroe2008fightin}.}
\label{tab:resistance_examples}
\end{table*}


Our resistance taxonomy comprises four coarse-grained behavioral patterns encompassing 13 fine-grained categories, offering a comprehensive framework for understanding client resistance in text-based counseling interactions. We present the framework from coarse- to fine-grained levels, and illustrate each category using a shared conversational context and counselor's intervention (see Table~\ref{tab:resistance_examples}): \textit{a client describes frequent arguments with her husband, and the counselor suggests choosing an appropriate time for a calm and constructive discussion}.

\noindent\textbf{Arguing.} Arguing a readily observable form of client resistance, often signaling interpersonal tension within the therapeutic relationship. Clients may become skeptical of the reliability of the counselor's statements or methods, actively disputing the counselor's interpretations and suggestions (\textit{\textbf{Challenging}}). More intensively, they may directly question the counselor's qualifications or expertise, implying that the counselor lacks sufficient understanding of their own intervention and thereby dismissing or devaluing the entire counseling process (\textit{\textbf{Discounting}}).

\noindent\textbf{Denying.} When faced with the counselor's invitation to explore potential directions for change, clients may display behaviors indicating misalignment with the counselor's guidance. Clients may outright reject the proposed approach without offering constructive alternatives (\textit{\textbf{Disagreeing}}). Resistance also emerges when clients deflect responsibility for change by attributing the issue to external factors. Clients may shift blame onto others (\textit{\textbf{Blaming}}), or point to objective circumstances that hinder their ability to follow the counselor's advice (\textit{\textbf{Excusing}}). Clients may also express defeat or hopelessness, conveying the belief that meaningful change is unattainable (\textit{\textbf{Pessimism}}). In addition, clients may demonstrate a lack of readiness to embrace change. They might minimize the severity of the issue or downplay the risks highlighted by the counselor, suggesting that no change is necessary (\textit{\textbf{Minimizing}}), or they may express outright unwillingness to try new approaches (\textit{\textbf{Unwillingness}}), or exhibit hesitation and uncertainty about taking action (\textit{\textbf{Reluctance}}).

\noindent\textbf{Avoiding.} When the counselor attempts to explore the client's inner emotions or past experiences more deeply, the client may exhibit behaviors that impede the therapeutic process. They might respond with brief, superficial remarks that lack substantive content (\textit{\textbf{Minimum Talk}}), or present various justifications to distance themselves from engaging with the topic (\textit{\textbf{Limit Setting}}).

\noindent\textbf{Ignoring.} Clients may show little attunement to the counselor's interventions, as if the two were moving along parallel tracks. Clients may remain absorbed in their own narrative, deeply immersed in emotional outpouring, without acknowledging the counselor's attempts (\textit{\textbf{Inattention}}). Alternatively, clients may abruptly introduce a new topic or concern, diverting the conversation away from the counselor's intended focus (\textit{\textbf{Sidetracking}}).

\section{Data Collection}
\label{sec: data}
To enable automated resistance detection, we curated the \textbf{ClientResistance} dataset following the PsyFIRE taxonomy.

\subsection{Data Source}
The counseling data utilized in this study originate from a restricted-access research dataset \textbf{ClientBehaviors}~\citep{li-etal-2023-understanding}. This dataset was collected from a Chinese online text-based counseling platform, where licensed counselors provide 50-minute sessions to real-world mental health seekers. It represents the largest real-world counseling conversation corpus in Mandarin, comprising 2,382 sessions with 93,121 client utterances. 



\subsection{Annotation Process}

During the formal annotation phase, annotators were assigned two tasks: (1) independently labeling utterances according to the proposed taxonomy, and (2) providing context-specific explanations to support their annotations. Further details are provided in Appendix~\ref{appendix: annotation}.

\noindent\textbf{Annotators Recruitment and Training.} Accurately identifying resistance behaviors presents a considerable challenge~\citep{lee-etal-2019-identifying, li-etal-2023-understanding} and demands specialized training~\citep{hara2015therapist}. To ensure annotation quality, we recruited and trained four annotators based on our resistance taxonomy. All annotators possessed academic qualifications in applied psychology or counseling psychology at the master's level. Before the formal labeling process, each annotator was required to complete qualifying exams and undertook more than 3 days of in-person training, which included detailed manual feedback on over 500 sample cases.

\begin{table}[]
\centering
\scalebox{0.63}{
\begin{tabular}{lcc|lcc}
\hline
\textbf{Categories} & \textbf{Num} & \textbf{Avg. Len} & \textbf{Categories}    & \textbf{Num} & \textbf{Avg. Len} \\ \hline
Challenging         & 682          & 26.03             & Reluctance             & 586          & 16.37             \\
Discounting         & 1094         & 33.18             & Minimum Talk           & 1018         & 6.41              \\
Disagreeing         & 1698         & 28.87             & Limit Setting          & 850          & 18.05             \\
Blaming             & 422          & 35.99             & Inattention            & 1040         & 32.01             \\
Excusing            & 336          & 29.10             & Sidetracking           & 1064         & 27.38             \\ 
Pessimism           & 856          & 32.15             & \textit{Resistance}    & \textit{10,398}       & \textit{25.53}             \\
Minimizing          & 368          & 22.59             & \textit{Collaboration} & \textit{13,532}       & \textit{31.79}             \\
Unwillingness       & 384          & 20.26             & \textit{\textbf{Total}}         & \textit{\textbf{23,930}}     & \textbf{\textit{29.02}}             \\ \hline
\end{tabular}}
\caption{Summary statistics of \textit{ClientResistance} dataset.}
\label{tab:data_statistics}
\end{table}

\noindent\textbf{Annotating Resistance.} In the annotation task, annotators were presented with conversation snippets that included the dialogue history between the counselor and the client, ending with the client's most recent response. Based on this context, annotators were instructed to categorize the client's final utterance into one of the 13 fine-grained resistance behavior types defined in our proposed taxonomy, or assign a \textit{Collaboration} label. Each sample was randomly assigned to two annotators for independent annotation. Samples where the annotators disagreed were reassigned to additional annotators until a final label was determined via majority vote. This process resulted in 23,930 annotated client utterances, with 10,398 classified as \textit{Resistance} behaviors.

\noindent\textbf{Crafting Explanations.} Following the categorical annotations, annotators were instructed to write context-sensitive explanations for each labeled sample. These explanations integrate linguistic rationales derived from the client's utterances with a conceptual interpretation of the observed resistant behavior in response to the counselor's intervention. For instance, in the \textit{Blaming} example provided in Table~\ref{tab:resistance_examples}, the explanation states: \textit{"The counselor suggests a calm discussion with the husband, to which the client responds by labeling him as `always so emotional' and insisting that `he's the one' who must improve. This response exemplifies a denial of the counselor's proposal through blame, externalizing responsibility entirely onto the spouse."}

\noindent\textbf{Quality Control.}
Inter-annotator agreement for resistance labels, measured by Cohen's $\kappa$, reached 0.72 overall, with class-level $\kappa$ ranging from 0.62 to 0.80, indicating substantial to near-perfect agreement~\citep{landis1977measurement}. To assess rationale quality, 200 samples were randomly selected and independently rated by two taxonomy developers on a 5-point Likert scale for \textit{Label Alignment}, \textit{Context Relevance}, \textit{Rationality}, and \textit{Clarity}, yielding consistently high scores ($3.32~\sim 3.80$). This evaluation demonstrates that both label assignments and rationales are reliable.

\section{RECAP: Model for Resistance Capture}
\label{sec: model}
Leveraging our collected dataset, we introduced \textbf{RECAP}, an automatic LLM-based framework comprising models for distinguishing resistance from collaboration and identifying fine-grained resistance types expressed in client utterances, while generating context-sensitive explanations.

\begin{table*}[]
\centering
\scalebox{0.465}{
\begin{tabular}{ll|cccc||ccc|ccc|cccc}
\hline\hline
& \multirow{2}{*}{\textbf{Model Name}} & \multicolumn{4}{c}{\textbf{Fine-grained Overall}}                                                     & \multicolumn{3}{c}{\textbf{Collaboration}}                                  & \multicolumn{3}{c}{\textbf{Resistance}}                                     & \multicolumn{4}{c}{\textbf{Binary Overall}}                                                           \\
&                                      & \textbf{P.}      & \textbf{R.}         & \textbf{F1}             & \textbf{Acc.}            & \textbf{P.}      & \textbf{R.}         & \textbf{F1}             & \textbf{P.}      & \textbf{R.}         & \textbf{F1}             & \textbf{P.}      & \textbf{R.}         & \textbf{F1}             & \textbf{Acc.}           \\ \hline
\multirow{8}{*}{\rotatebox{90}{Zero-Shot}} & GPT-4o                               & $48.01_{0.98}$          & $49.74_{1.23}$          & $45.22_{0.63}$          & $47.62_{0.50}$          & $76.66_{0.77}$          & $93.70_{0.39}$          & $84.33_{0.52}$          & $88.47_{0.70}$          & $62.86_{1.57}$          & $73.49_{1.21}$          & $82.56_{0.65}$          & $78.28_{0.83}$          & $78.91_{0.86}$          & $80.30_{0.74}$          \\
                           & Claude-3.5-Sonnet                    & $49.69_{1.17}$          & $46.19_{1.03}$          & $43.65_{1.52}$          & $46.87_{1.22}$          & $76.84_{0.65}$          & $95.58_{0.59}$          & $85.19_{0.54}$          & \bm{$91.58_{1.10}$} & $62.51_{1.28}$          & $74.30_{1.10}$          & $84.21_{0.78}$          & $79.05_{0.77}$          & $79.75_{0.82}$          & $81.21_{0.72}$          \\
                           & Qwen2.5-7B-Instruct                  & $33.39_{1.91}$          & $23.08_{5.60}$          & $21.21_{5.60}$          & $23.76_{5.43}$          & $68.87_{0.70}$          & $95.99_{0.69}$          & $80.20_{0.48}$          & $89.33_{1.53}$          & $43.51_{1.99}$          & $58.49_{1.79}$          & $79.10_{0.84}$          & $69.75_{0.90}$          & $69.34_{1.11}$          & $73.19_{0.78}$          \\
                           & Qwen2.5-14B-Instruct                 & $39.57_{2.70}$          & $31.10_{1.99}$          & $25.06_{1.97}$          & $27.81_{2.46}$          & $73.43_{1.83}$          & $89.30_{1.56}$          & $79.93_{0.62}$          & $83.51_{1.00}$          & $56.18_{1.24}$          & $65.23_{1.70}$          & $78.47_{0.91}$          & $72.74_{0.84}$          & $72.58_{1.15}$          & $74.91_{1.40}$          \\
                           & Qwen2.5-32B-Instruct                 & $37.64_{1.85}$          & $39.01_{2.48}$          & $30.76_{1.33}$          & $32.97_{1.31}$          & $74.77_{0.52}$          & $87.09_{1.29}$          & $79.73_{0.30}$          & $81.51_{1.06}$          & $59.90_{1.52}$          & $67.01_{0.43}$          & $78.14_{0.27}$          & $73.50_{0.31}$          & $73.37_{0.90}$          & $75.28_{0.90}$          \\
                           & Qwen2.5-72B-Instruct                 & $40.44_{1.41}$          & $40.51_{1.22}$          & $36.44_{0.77}$          & $39.49_{0.94}$          & $76.54_{0.80}$          & $85.18_{0.42}$          & $80.63_{0.47}$          & $77.38_{0.59}$          & $66.01_{1.55}$          & $71.24_{1.06}$          & $76.96_{0.63}$          & $75.59_{0.76}$          & $75.93_{0.76}$          & $76.85_{0.67}$          \\
                           & Llama-3.1-8B-Instruct                & $29.27_{0.81}$          & $28.98_{1.02}$          & $26.19_{1.30}$          & $28.83_{1.09}$          & $62.82_{1.30}$          & $85.35_{1.32}$          & $72.07_{1.17}$          & $84.76_{1.26}$          & $60.27_{1.54}$          & $69.85_{0.76}$          & $73.79_{1.07}$          & $72.81_{0.62}$          & $70.96_{0.96}$          & $71.17_{0.78}$          \\
                           & Llama-3.1-70B-Instruct               & $47.61_{2.05}$          & $35.76_{3.11}$          & $32.91_{3.30}$          & $34.64_{3.53}$          & $77.78_{0.97}$          & $83.71_{1.18}$          & $80.64_{1.01}$          & $76.47_{1.57}$          & $68.88_{1.44}$          & $72.47_{1.41}$          & $77.13_{1.23}$          & $76.30_{1.18}$          & $76.56_{1.20}$          & $77.27_{1.17}$          \\ \hline
\multirow{8}{*}{\rotatebox{90}{Few-Shot}}  & GPT-4o                               & $49.00_{1.39}$          & $49.79_{0.86}$          & $45.37_{0.82}$          & $47.84_{0.94}$          & $77.96_{1.14}$          & $94.85_{1.16}$          & $85.57_{0.48}$          & $90.71_{1.62}$          & $65.06_{2.63}$          & $75.73_{1.43}$          & $84.34_{0.62}$          & $79.96_{0.92}$          & $80.65_{0.92}$          & $81.91_{0.74}$          \\
                           & Claude-3.5-Sonnet                    & $49.45_{1.12}$          & $46.24_{1.72}$          & $44.99_{1.43}$          & $47.89_{1.39}$          & $80.20_{0.78}$          & $91.26_{1.06}$          & $85.37_{0.47}$          & $86.16_{1.29}$          & $70.67_{1.62}$          & $77.63_{0.85}$          & $83.18_{0.64}$          & $80.96_{0.63}$          & $81.50_{0.62}$          & $82.31_{0.57}$          \\
                           & Qwen2.5-7B-Instruct                  & $29.12_{1.92}$          & $28.01_{1.36}$          & $24.25_{0.89}$          & $27.75_{1.42}$          & $69.43_{0.51}$          & \bm{$96.12_{0.40}$} & $80.63_{0.47}$          & $89.91_{1.16}$          & $44.95_{1.11}$          & $59.94_{1.24}$          & $79.67_{0.83}$          & $70.54_{0.74}$          & $70.28_{0.86}$          & $73.88_{0.69}$          \\
                           & Qwen2.5-14B-Instruct                 & $38.08_{2.40}$          & $32.54_{1.37}$          & $25.92_{1.02}$          & $30.49_{0.81}$          & $73.82_{1.49}$          & $88.54_{1.07}$          & $79.97_{0.81}$          & $82.24_{1.31}$          & $57.67_{1.70}$          & $66.14_{1.36}$          & $78.03_{1.00}$          & $73.11_{1.13}$          & $73.05_{1.26}$          & $75.13_{1.17}$          \\
                           & Qwen2.5-32B-Instruct                 & $38.48_{2.97}$          & $40.92_{2.37}$          & $32.76_{1.69}$          & $34.99_{1.72}$          & $76.74_{0.51}$          & $83.49_{1.37}$          & $79.53_{0.33}$          & $77.12_{1.15}$          & $66.15_{1.54}$          & $70.27_{0.40}$          & $76.93_{0.72}$          & $74.82_{0.89}$          & $74.90_{0.83}$          & $75.96_{0.70}$          \\
                           & Qwen2.5-72B-Instruct                 & $39.28_{1.11}$          & $41.42_{1.26}$          & $36.47_{0.96}$          & $40.20_{0.47}$          & $81.44_{2.58}$          & $78.39_{3.87}$          & $79.76_{0.91}$          & $73.32_{2.18}$          & $76.44_{5.06}$          & $74.68_{1.38}$          & $77.38_{0.39}$          & $77.42_{0.76}$          & $77.22_{0.55}$          & $77.55_{0.46}$          \\
                           & Llama-3.1-8B-Instruct                & $40.65_{1.90}$          & $27.02_{1.97}$          & $27.14_{2.01}$          & $29.54_{2.28}$          & $74.59_{1.94}$          & $85.38_{1.39}$          & $78.49_{0.83}$          & $80.11_{1.85}$          & $59.56_{1.07}$          & $65.72_{1.18}$          & $77.35_{0.66}$          & $72.47_{1.10}$          & $72.11_{1.26}$          & $74.16_{1.09}$          \\
                           & Llama-3.1-70B-Instruct               & $47.69_{1.16}$          & $41.71_{1.61}$          & $39.25_{1.84}$          & $41.24_{1.78}$          & $81.27_{2.29}$          & $79.78_{3.97}$          & $80.42_{1.23}$          & $74.43_{2.53}$          & $75.86_{4.57}$          & $75.01_{1.29}$          & $77.85_{0.73}$          & $77.82_{0.86}$          & $77.72_{0.83}$          & $78.08_{0.84}$          \\ \hline
& \textbf{Our Model}                   & \bm{$67.02_{1.37}$} & \bm{$66.76_{1.41}$} & \bm{$66.58_{1.30}$} & \bm{$67.72_{0.99}$} & \bm{$92.14_{0.93}$} & $92.73_{0.88}$          & \bm{$92.43_{0.39}$} & $90.47_{0.97}$          & \bm{$89.69_{1.39}$} & \bm{$90.07_{0.55}$} & \bm{$91.31_{0.44}$} & \bm{$91.21_{0.51}$} & \bm{$91.25_{0.46}$} & \bm{$91.41_{0.45}$} \\ \hline\hline
\end{tabular}}
\caption{Classification performance on binary and fine-grained resistance behavior tasks for baseline models and our model. Best results in each column are shown in \textbf{bold}.}
\label{tab:main_classification_results}
\end{table*}

\begin{table*}[]
\centering
\scalebox{0.465}{
\begin{tabular}{lcc||cccc||ccc|ccc|cccc}
\hline\hline
\multirow{2}{*}{\textbf{Backbone}}     & \multirow{2}{*}{\textbf{Rationale}} & \multirow{2}{*}{\textbf{Fine-tune}} & \multicolumn{4}{c}{\textbf{Fine-grained Overall}}                 & \multicolumn{3}{c}{\textbf{Collaboration}}       & \multicolumn{3}{c}{\textbf{Resistance}}          & \multicolumn{4}{c}{\textbf{Binary Overall}}                       \\
                                       &                                     &                                     & \textbf{P.}    & \textbf{R.}    & \textbf{F1}    & \textbf{Acc.}  & \textbf{P.}    & \textbf{R.}    & \textbf{F1}    & \textbf{P.}    & \textbf{R.}    & \textbf{F1}    & \textbf{P.}    & \textbf{R.}    & \textbf{F1}    & \textbf{Acc.}  \\ \hline
\multirow{4}{*}{Llama-3.1-8B-Instruct} & \ding{51}         & FPFT                                & $67.02_{1.37}$ & $66.76_{1.41}$ & $66.58_{1.30}$ & $67.72_{0.99}$ & $92.14_{0.93}$ & $92.73_{0.88}$ & $92.43_{0.39}$ & $90.47_{0.97}$ & $89.69_{1.39}$ & $90.07_{0.55}$ & $91.31_{0.44}$ & $91.21_{0.51}$ & $91.25_{0.46}$ & $91.41_{0.45}$ \\
                                       & \ding{55}          & FPFT                                & $62.44_{1.40}$ & $61.42_{1.48}$ & $61.72_{1.52}$ & $63.36_{1.76}$ & $91.12_{4.86}$ & $88.74_{6.48}$ & $89.68_{0.99}$ & $86.45_{6.63}$ & $88.17_{8.09}$ & $86.89_{1.18}$ & $88.79_{1.07}$ & $88.46_{1.05}$ & $88.29_{0.68}$ & $88.49_{0.68}$ \\
                                       & \ding{51}          & LoRA                                & $54.81_{1.97}$ & $50.30_{1.62}$ & $49.54_{1.19}$ & $51.42_{1.09}$ & $81.72_{0.88}$ & $86.45_{0.81}$ & $84.01_{0.78}$ & $80.92_{1.10}$ & $74.82_{1.33}$ & $77.75_{1.15}$ & $81.32_{0.95}$ & $80.63_{0.96}$ & $80.88_{0.96}$ & $81.40_{0.93}$ \\
                                       &  \ding{55}                                   & \ding{55}                                  & $29.27_{0.81}$ & $28.98_{1.02}$ & $26.19_{1.30}$ & $28.83_{1.09}$ & $77.78_{0.97}$ & $83.71_{1.18}$ & $80.64_{1.01}$ & $76.47_{1.57}$ & $68.88_{1.44}$ & $72.47_{1.41}$ & $77.13_{1.23}$ & $76.30_{1.18}$ & $76.56_{1.20}$ & $77.27_{1.17}$ \\ \hline
\multirow{4}{*}{Qwen2.5-7B-Instruct}   &    \ding{51}                                 & FPFT                                & $63.90_{1.34}$ & $62.74_{1.50}$ & $62.91_{1.11}$ & $64.40_{1.46}$ & $91.45_{1.14}$ & $91.65_{1.06}$ & $91.54_{0.38}$ & $89.12_{1.08}$ & $88.82_{1.74}$ & $88.96_{0.61}$ & $90.29_{0.44}$ & $90.24_{0.55}$ & $90.25_{0.48}$ & $90.42_{0.46}$ \\
                                       &   \ding{55}                                  & FPFT                                & $61.63_{2.80}$ & $59.41_{1.29}$ & $59.49_{1.77}$ & $61.15_{2.22}$ & $87.36_{1.35}$ & $89.26_{2.42}$ & $88.27_{0.98}$ & $85.73_{2.62}$ & $83.14_{2.27}$ & $84.36_{1.08}$ & $86.54_{1.18}$ & $86.20_{0.94}$ & $86.32_{1.00}$ & $86.60_{1.01}$ \\
                                       &    \ding{51}                                 & LoRA                                & $51.60_{1.55}$ & $45.34_{1.29}$ & $45.09_{0.91}$ & $46.95_{1.00}$ & $81.11_{0.95}$ & $85.15_{0.68}$ & $83.08_{0.82}$ & $79.33_{1.04}$ & $74.19_{1.40}$ & $76.67_{1.23}$ & $80.22_{0.99}$ & $79.67_{1.03}$ & $79.87_{1.02}$ & $80.38_{0.98}$ \\
                                       &     \ding{55}                                & \ding{55}                                  & $33.39_{1.91}$ & $23.08_{5.60}$ & $21.21_{5.60}$ & $23.76_{5.43}$ & $76.54_{0.80}$ & $85.18_{0.42}$ & $80.63_{0.47}$ & $77.38_{0.59}$ & $66.01_{1.55}$ & $71.24_{1.06}$ & $76.96_{0.63}$ & $75.59_{0.76}$ & $75.93_{0.76}$ & $76.85_{0.67}$ \\ \hline\hline
\end{tabular}}
\caption{Ablation study results on the effects of rationale inclusion and training strategies on Llama-3.1-8B-Instruct and Qwen2.5-7B-Instruct.}
\label{tab:ablation}
\end{table*}

\subsection{Problem Definition}
\label{problem_definition}
Let $H_i$ denote the conversation history, ending with the therapist's intervention, and $R_i$ the client's response. Each pair ($H_i$, $R_i$) is annotated as either Collaboration or one of 13 fine-grained Resistance subtypes. Based on this annotation, we define two tasks: (1) \textbf{binary classification} to distinguish \textit{Collaboration} from \textit{Resistance}, and (2) \textbf{fine-grained classification} to identify the specific resistance subtype for instances labeled as \textit{Resistance}. In both tasks, models additionally generate free-text explanatory rationales for their predictions.


\subsection{Experimental Setup}

\noindent\textbf{Cross-Validation and Data Splitting.} To ensure a robust and unbiased performance estimate, we employed a stratified 5-fold cross-validation strategy. The dataset was partitioned into five mutually exclusive folds using stratified random sampling, preserving the original class distribution in each fold. This design mitigates bias from random splits and provides a stable evaluation.

\noindent\textbf{Training Configuration.}
For each fold in the cross-validation procedure, we fine-tuned the LLaMA-3.1-8B-Instruct model using supervised full-parameter fine-tuning (FPFT), with four folds serving as the training set and the remaining fold as the validation set. The training objective minimized the cross-entropy loss over both behavior classification and rationale generation. To control overfitting, training was capped at 10 epochs with early stopping based on validation loss, and the model checkpoint yielding the lowest validation loss was used for final evaluation.

\noindent\textbf{Inference and Evaluation.}
During inference, the sampling temperature was set to 0 and top-$p$ to 1.0 to ensure deterministic outputs. Model performance was reported as the average with standard deviation of the evaluation metric across all five validation folds, reflecting both average performance and variability across data splits.




\subsection{Baselines}
We further assessed a variety of advanced LLMs under both zero-shot and few-shot prompting settings. The evaluation covers leading closed-source models, including GPT-4o~\citep{2023gpt4} and Claude-3.5-Sonnet~\citep{Anthropic2024}, as well as open-source models from the Qwen2.5-Instruct family (7B, 14B, 32B, and 72B)~\citep{qwen2.5}, the Llama-3.1-Instruct family (8B and 70B)~\citep{llama3modelcard}. All baseline models were evaluated using the same protocol as RECAP to ensure fair comparison. Further implementation details are provided in Appendix~\ref{appendix:implementation}.


\section{Experiment Results}
\label{sec: results}
In this section, we analyze the effectiveness of our computational approach in identifying resistance and generating underlying rationales.

\subsection{Main Results}


Table~\ref{tab:main_classification_results} reports precision, recall, macro-F1, and accuracy for all models on both binary and fine-grained classification tasks, with per-category results provided in Appendix~\ref{appendix: results}. 

Overall, \textbf{general-purpose LLMs struggle to accurately detect client resistance without domain-specific demonstrations.} Across models, F1 scores for \textit{Collaboration} are substantially higher than those for \textit{Resistance}, with \textit{Resistance} F1 trailing by roughly 10 points. This gap is primarily attributable to low recall on resistance instances, indicating that general LLMs tend to overlook resistance signals. The challenge is more pronounced for fine-grained resistance categories, where both F1 and accuracy generally remaining < 50\%, and in some cases dropping < 30\% (e.g., Qwen2.5-Instruct 7B and 14B, and LLaMA-3.1-8B-Instruct). Most errors occur in the \textit{Inattention} category, underscoring the models' weakness in recognizing indirect or implicit expressed resistance.

\textbf{Incorporating few-shot demonstrations leads to consistent performance gains across nearly all baselines}, demonstrating the value of our task‑tailored dataset. With in-context examples, GPT-4o and Claude-3.5-Sonnet achieve the strongest results, reaching around 82\% F1 in binary classification and 45\% F1 in fine-grained classification. Nevertheless, these performance ceilings remain limited, reaffirming the inherent difficulty of reliably detecting and differentiating resistance from textual utterances.

\textbf{Our RECAP model substantially outperforms all baselines}. In binary classification, RECAP achieves over 91\% F1 and accuracy representing gains of +9.75 F1 and +9.10 accuracy over best baselines. These improvements are largely driven by a marked increase in recall for \textit{Resistance}, reaching nearly 90\% while maintaining high precision. This balanced performance results narrow the gap between resistance and collaboration to within 2 F1 points (v.s 10+ points for baselines). RECAP also demonstrates robustness in fine-grained resistance classification, outperforming the strongest baseline by +21.21 in F1 and +19.83 in accuracy. This capability represents a critical advancement for real-world deployment, enabling precise detection of subtle and varied resistance patterns essential for effective conversational analysis and intervention.

\subsection{Ablation Study}

We perform ablation studies to assess the impact of our annotated dataset and rationale-augmented training on model performance. We compare different training strategies (FPFT vs. LoRA~\citep{lora}, rank 8) and replicate experiments on Qwen2.5-7B-Instruct to assess generalizability.

As shown in Table~\ref{tab:ablation}, \textbf{performance gains arise mainly from training on our task-specific annotated dataset}. LoRA training alone already provides substantial improvements, roughly doubling F1 in the challenging fine-grained resistance classification task. FPFT delivers further gains, improving F1 by about +10 in binary classification and +15 in fine-grained classification. Compared with training on behavior classification alone, incorporating rationales boosts F1 across all resistance subcategories. This indicate that integrating explanations as auxiliary information can further enhance the model's ability to handle challenging classification cases.


\subsection{Error Analysis}

We qualitatively analyzed RECAP's classification errors. In distinguishing \textit{Resistance} from \textit{Collaboration}, we found that the model sometimes failed to identify subtle expressions of resistance, especially when the expression lacks overt linguistic forms of negation. For instance, when a counselor recommended meditation as a widely endorsed method for stress relief, a client's response such as "Maybe only the methods that suit me will work" was often incorrectly interpreted as collaborative. These observations underscore that accurately identifying resistance requires more than detecting surface-level linguistic cues; it demands a deeper contextual understanding of the function of client responses. 

In the fine-grained resistance classification task, we observed that emotionally charged expressions of hopelessness were sometimes misinterpreted as rational disagreement. For instance, when a counselor encouraged action, clients' statements like I think my life is basically over" were often labeled as disagreement. In reality, such statements reflect emotional withdrawal rather than a reasoned objection. This highlights the importance of differentiating between affect-driven and cognition-driven forms of resistance, a distinction that remains challenging for current models.


\section{Model-based Insights into Online Counseling}
\label{sec: application}

We applied RECAP to the CounselingWAI dataset~\citep{li-etal-2024-understanding-therapeutic}, which originates from a study involving text-based online counseling between clients and licensed professionals. Our analysis focused on 793 sessions with client-reported working alliance scores, aiming to examine (1) the prevalence and patterns of resistance, and (2) its impact on therapeutic alliance development. We further designed a proof-of-concept study to illustrate the potential advantages of real-time, automated resistance detection during counseling.

\paragraph{Prevalence and Diversity of Resistance.} We observed that resistance occurred in over 90\% of sessions, with about 17\% of client utterances exhibiting resistance, confirming it as a recurrent element of online counseling. The most common subtypes were \textit{Disagreeing} and \textit{Inattention}, underscoring the need to attend to both explicit disagreement and subtle disengagement. On average, clients exhibited more than two distinct fine-grained resistance types per session, emphasizing the diversity and complexity of resistance expressions.

\paragraph{Relationship Ruptures Are Associated with High Resistance Levels.} Extending prior psychotherapy research linking relationship ruptures to poorer outcomes and increased dropout rates~\citep{beutler2001resistance}, we analyzed the association between resistance and therapeutic alliance strength. Pearson correlation analysis revealed a significant negative correlation between session-level resistance and client-reported working alliance scores ($\rho = -0.22$, $p < 0.001$), indicating that higher resistance is associated with a weaker therapeutic alliance. Notably, client \textit{discounting} exhibited the strongest negative correlation with overall alliance ($\rho = -0.26$, $p < 0.001$), particularly impacting the bond dimension ($\rho = -0.23$, $p < 0.001$). These findings suggest counselors prioritize trust repair and relationship restoration in such instances.

\paragraph{Implications for Model-based Feedback of Resistance.}

\begin{figure}
    \centering
    
    \includegraphics[width=1\linewidth]{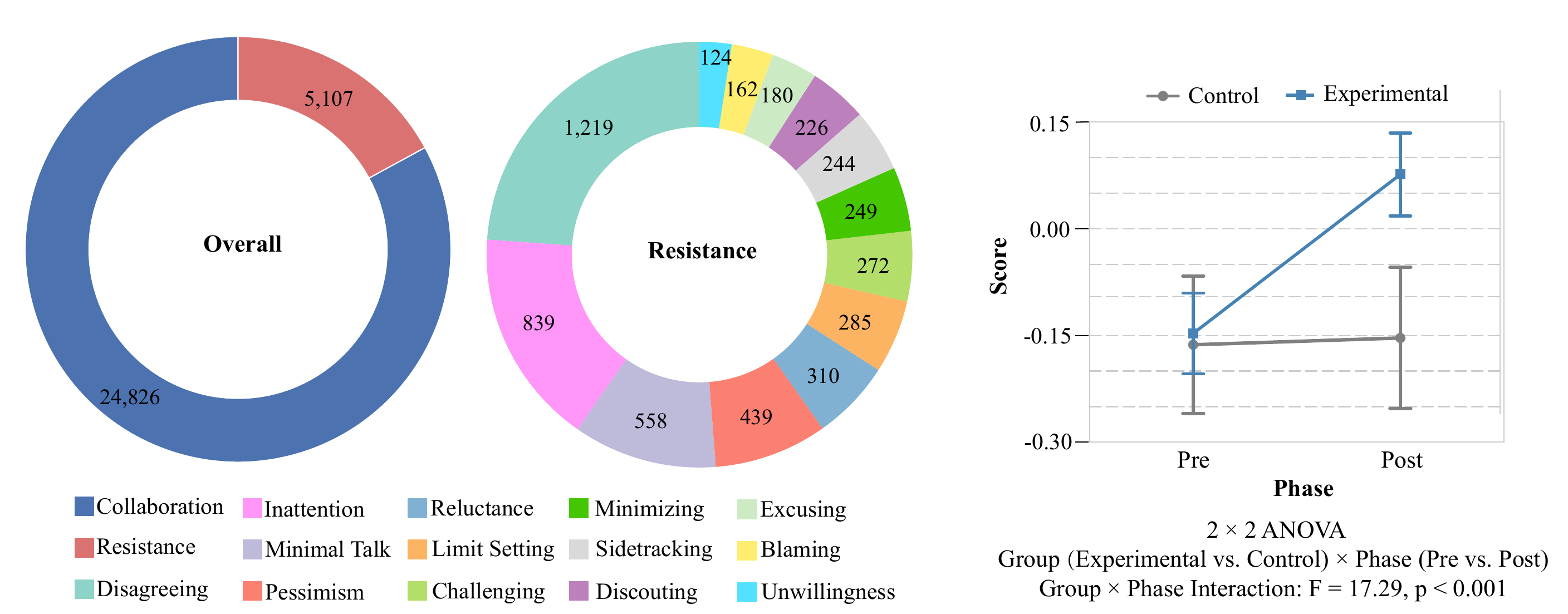}
    \caption{Left: Donut charts showing the distribution of collaboration versus resistance and fine-grained resistance types in the CounselingWAI dataset, as predicted by RECAP. Right: Pre- and post-test mean scores (±SD) for control and experimental groups, with ANOVA results from the proof-of-concept study.}
    \label{fig:poc_experiment}
\end{figure}

To examine whether model-based feedback can improve counselors' understanding of and responses to client resistance, we conducted a proof-of-concept study with 62 counselors using a mixed between- and within-subjects design. Participants were randomly assigned to a control or experimental group. All counselors first responded to 30 client resistance scenarios, and then revised their responses; only the experimental group received model-generated feedback about the resistance type and explanatory rationales.

Responses were evaluated using established resistance management principles on a three-point scale~\citep{rosengren2008video} ($-1$: increases resistance, $0$: neutral, and $1$: reduces resistance. A 2 (Group) $\times$ 2 (Phase) mixed-design ANOVA revealed a significant Group $\times$ Phase interaction ($F = 17.29$, $p < .001$, $\eta^2 = 0.24$). Follow-up analyses showed a substantial improvement in the experimental group (Cohen's $d = 1.17$), but no significant change in the control group ($d = 0.05$). An independent t-test at post-phase further confirmed the experimental group's superior performance ($d = 0.59$).


Participants in experimental group also rated the feedback as highly helpful for recognizing (mean = 4.47) and managing (mean = 4.19) client resistance. In interviews, counselors reported:~\textit{"When responding on my own, I sometimes missed signs of client resistance. The feedback helped me see when the conversation got stuck and understand why, more quickly and accurately."} They further noted:~\textit{"The detailed resistance behaviors, together with rationales, made it much clearer how to adjust my strategies and handle these critical moments appropriately."} Overall, these qualitative and quantitative findings suggest that the AI tool can help counselors detect misalignments with clients more promptly and responding more adaptively during critical moments in therapy.






\section{Conclusion}

We introduce a novel framework, dataset, and automated model RECAP for identifying client resistance in text-based mental health counseling conversations. RECAP not only accurately detects instances of resistance but also generates corresponding explanatory feedback. Applied at scale, the model reveals resistance patterns and highlights the role of resistance in shaping therapeutic relationship. We also demonstrate that offering model-based feedback could support counselors to better recognize and address resistance. Code, model, and annotated data are available at \href{https://anonymous.4open.science/r/ClientResistance-EMNLP/}{this URL}.

\section{Limitation}
As the first effort to automatically identify diverse expressions of client resistance in text-based counseling, this study provides a foundational step, though there remains considerable room for improvement. 

Regarding the resistance taxonomy, the proposed framework primarily focuses on interpersonal resistance, that is, client behaviors in response to the counselor's in-session interventions. It does not yet encompass other prevalent forms of resistance frequently observed in therapeutic settings, such as chronic lateness or persistent failure to complete homework assignments. Moreover, the taxonomy was developed and validated exclusively within Chinese counseling contexts. While certain expressions of resistance may exhibit cross-cultural consistency—allowing models trained on taxonomy-based annotations to achieve a degree of generalizability across languages and cultural settings—meaningful cultural variations still exist. To improve both the robustness and cross-context applicability of resistance identification models, future research should incorporate broader cross-cultural perspectives.
\section{Ethics Statement}

\paragraph{Data.} The psychological counseling datasets used in this study were collected in prior research and are available to the academic community upon formal request. All counseling dialogues were thoroughly anonymized by human reviewers to ensure compliance with ethical standards and the protection of participant confidentiality. Permission to use the data was obtained through a formal data-sharing agreement with the original data providers. With the approval of the data providers, we also plan to publicly release the annotated dataset constructed in this study to facilitate future research on client resistance. The released dataset will be made available for research purposes only and distributed under a data-sharing agreement that outlines ethical usage guidelines.

\paragraph{Ethics Approval.}
This study has been reviewed and approved by the Institutional Review Board (IRB) at [Institution anonymized for review], under protocol number [Anonymized for review].

\paragraph{Human Annotators.}
All annotators received training in psychological counseling principles and provided informed consent prior to participation. They were fully briefed on the goals of the study, the nature of the data, and the importance of maintaining confidentiality. Annotators were fairly compensated for their contributions, with each receiving an average of 2,000 RMB for their participation.

\paragraph{Model Deployment and Intended Use.}
The model developed in this study is intended strictly for academic research and experimental analysis. Its primary purpose is to advance the understanding of client resistance in text-based counseling and to support further investigations into human–AI collaboration in mental health contexts. It is not designed for clinical deployment or for making autonomous therapeutic decisions. The model is not a substitute for professional mental health judgment, and in any real-world scenario, its outputs should be interpreted cautiously.

\paragraph{Philosophical and Ethical Implications}

Client resistance is a natural and meaningful part of the therapeutic process. It often signals underlying emotional pain, internal conflict, or unmet psychological needs. We believe such moments should be met with curiosity and compassion—not control.

The essence of therapeutic counseling is not to override resistance or enforce compliance, but to foster a space where clients feel heard, respected, and empowered to pursue their goals at their own pace. In this work, we present a framework and automated model designed to help counselors identify critical moments of resistance—not to suppress them, but to support empathic, autonomy-affirming care. Our goal is to ensure that RECAP is never framed as a tool for eliminating resistance, but as one that helps counselors engage with it thoughtfully and respectfully.


\bibliography{custom}

\appendix
\section{Resistance Framework}
\subsection{Framework Development}
\label{appendix: framework_development}
We engaged three primary taxonomy developers to construct the framework, following the principles of consensual qualitative research~\citep{hill1997CQR}. The team comprised a postdoctoral researcher specializing in psychology and conversation analysis, a postgraduate student majoring in clinical psychology with hands-on counseling experience, and a PhD student with expertise in the interdisciplinary field of natural language processing (NLP) and psychological counseling.

Drawing on existing taxonomies designed for face-to-face therapy~\citep{chamberlain1984observation, miller1999enhancing, ribeiro2014TCCS}, the developer team first excluded categories unsuitable for text-only conversation settings (e.g., \textit{interrupt}, \textit{talk over}, \textit{silence}, etc.), and developed an initial version of the taxonomy alongside accompanying annotation guidelines. Subsequently, 400-450 client utterances with conversation history were randomly selected, and two of the developers independently annotated them by assigning the single most salient category which best captured the client's primary stance and offered the most useful guidance for shaping the counselor's next response~\citep{carletta-etal-1997-reliability, sajid2023single}. Following annotation, all three developers engaged in in-depth group discussions to resolve discrepancies. These discussions centered on reconciling inconsistent labels in the dependent annotations, clarifying ambiguous definitions, and refining the boundaries between closely related categories. Typical examples and challenging cases were also incorporated into the guidelines to further align the annotators' understanding (see Appendix~\ref{appendix: label_comparison}). During this iterative refinement, categories were added, removed or merged as necessary. These collaborative efforts substantially reduced subjective variability and helped standardize the annotation process, ensuring more consistent and reliable results.

This iterative process was repeated three times, resulting in the final version of the taxonomy and guidelines, which include clear definitions and representative examples. The Cohen's kappa values between the two annotators across the three iterations were 0.47, 0.60 and 0.71, showing a consistent increase in inter-annotator agreement. This upward trend demonstrates the effectiveness of the iterative refinement process in resolving ambiguities and ensuring alignment among developers. The final level of agreement indicates substantial reliability of the taxonomy.

In total, 1,291 client utterances were annotated during the development phase of the taxonomy and guidelines.

\subsection{Definition and Examples of Client Resistance}
\label{appendix: resistance_definition}
\paragraph{Arguing.} The client challenges the counselor's professionalism or integrity, questions their qualifications or experience, expresses skepticism about the appropriateness of their interventions, or conveys dissatisfaction by implying that the counselor does not truly understand their situation:

\indent\textbf{Challenging.} The client directly questions the accuracy or truthfulness of the information or content provided by the counselor. (e.g. \textit{So what are you going to do about it, cure me?})

\indent\textbf{Discounting.} The client questions the counselor's personal abilities, professional knowledge, or role in the counseling process. (e.g. \textit{You clearly don't know what you're doing. This is a waste of time.})

\paragraph{Denying.} The client expresses unwillingness or inability to recognize problems, cooperate, accept responsibility, or take advice:

\indent\textbf{Blaming.} The client expresses resistance by deflecting responsibility and attributing the issue to others, thus refusing to align with the counselor's approach, particularly when the counselor offers suggestions for change. (e.g. \textit{It's my husband who's always so emotional; he's the one who really needs to work on his communication skills.})

\indent\textbf{Disagreeing.} When the counselor employs directive strategies—such as encouraging reflection, helping the client view the problem from a new perspective, suggesting possible solutions, or offering direct advice—the client demonstrates resistance by expressing disagreement without offering any constructive alternatives. (e.g. \textit{I think this approach might not work for me.})

\indent\textbf{Excusing.} The client conveys their inability to follow the counselor's guidance by offering excuses for their behavior and attributing the issue to external, objective factors. (e.g. \textit{Work has been really hectic lately. I probably won't be able to keep up with daily meditation.})

\indent\textbf{Minimizing.} The client implies that the counselor has exaggerated the risks or dangers, downplaying the situation as less severe than suggested, and conveying that their issues are not serious enough to warrant the counselor's guidance or assistance. (e.g. \textit{I don't think it's as serious as you're making it out to be.})

\indent\textbf{Pessimism.} The client expresses pessimistic, defeatist, or negative views about themselves, signaling their inability to follow the counselor's suggested course of action. (e.g. \textit{It's pointless—my life isn't going to get any better anyway.})

\indent\textbf{Reluctance.} The client shows hesitation or expresses reservations toward the information or suggestions offered by the counselor. (e.g. \textit{I understand what I should do, but I need to think it over.})

\indent\textbf{Unwillingness.} The client expresses contentment with the current situation and shows a lack of willingness to change, or conveys a resistance to making any changes. (e.g. \textit{I just don't want to try that method.})

\paragraph{Avoiding.} The client provides brief responses or withholds replies to minimize communication with the counselor on the current topic, thereby avoiding deeper self-exploration and problem-solving:

\indent\textbf{Minimal Talk.} The client withholds information, providing brief and superficial responses that lack depth and sufficient detail, often conveying a dismissive attitude. (e.g. \textit{It was just some complicated stuff.})

\indent\textbf{Limit Setting.} The client explicitly refuses or sidesteps discussing certain topics, often providing alternative reasons for doing so. (e.g. \textit{I don't really remember how I felt back then.})

\paragraph{Ignoring.} The client shows signs of neglect or non-compliance with the counselor's guidance. The client's responses often suggest that they have not followed the counselor's approaches. It may appear as though the client has completely disregarded the counselor's words or is acting as if the counselor's statements or questions were never made: 

\indent\textbf{Inattention.} The client remains fixated on their previously chosen topic, deeply immersed in the original subject or emotions, disregarding the counselor's attempts to intervene. (e.g. \textit{Oh right, something else happened a few days ago... (while the counselor is still exploring a previously mentioned issue)})

\indent\textbf{Sidetracking.} The client shifts the conversation away from the counselor's intended focus, introducing new topics or areas of attention. (e.g. \textit{My roommate forgot to bring me lunch again yesterday... (even as the counselor has initiated a breathing exercise)})

\subsection{Label Comparison}
\label{appendix: label_comparison}

To support annotators in making more precise distinctions, we have curated a set of representative situational examples in the guidelines to illustrate subtle differences between labels. Shown here is a typical case that distinguishing \textit{Excusing} from a client's sincere explanation of external circumstances that hinder therapeutic progress. 

Excusing is characterized by a subtle deflection of personal responsibility or a reduced motivation to engage in change. These expressions often rely on vague or generalized justifications (e.g., "I'm usually too busy to practice mindfulness") and may include frequent use of qualifiers such as "but" or "however," which implicitly downplay the need for emotional or behavioral engagement. The underlying intent is often to avoid deeper therapeutic work or to passively resist the therapeutic direction.

In contrast, when clients describe genuine external constraints, their explanations tend to be concrete, context-specific, and temporally bounded (e.g., "I probably can't do the mindfulness exercise tonight because of an unexpected overtime assignment, but I should be able to resume tomorrow"). These statements typically convey a tone of regret or frustration, and are often followed by indications of willingness to collaborate on alternative plans or adjustments. In such cases, the client's primary intent is to communicate authentic situational barriers and seek the counselor's understanding or adaptive support, rather than to impede therapeutic progress.

\section{Data Annotation}
\label{appendix: annotation}

\subsection{Annotators Recruitment and Training.}
We recruited undergraduate and graduate-level students in counseling or applied psychology from a variety of institutions to participate in the annotation task. As part of the selection process, all candidates were required to take offline exams. Initially, each candidate familiarized themselves with the annotation guidelines before completing three exams. Each exam consisted of approximately fifty conversation snippets, covering all 13 resistance categories and collaboration behaviors. For each snippet, candidates were tasked with annotating the client's final response to the counselor's intervention.

After each exam, we provided candidates with feedback on any misclassified annotations, along with the corresponding correct labels, to help them better understand the guidelines. The top four candidates with the highest average accuracy across all three exams, and who achieved at least a 60\% accuracy rate on the final exam, were selected as the final annotators (refer to Table~\ref{tab:annotator_exam}).

These selected annotators then underwent three days of offline training. During the training, all annotators annotated over 500 utterances simultaneously, with ground-truth labels provided by our psychological experts. Following the annotation process, group meetings and one-on-one consultations with our taxonomy developers were held to help annotators analyze and correct any mislabeled utterances.

Annotations were conducted using Doccano\footnote{https://github.com/doccano/doccano}, an open-source text annotation platform. 

\begin{table}[]
\centering
\scalebox{0.85}{
\begin{tabular}{ccccc}
\toprule
\textbf{ID} & \textbf{Exam 1} & \textbf{Exam 2} & \textbf{Exam 3} & \textbf{Avg.} \\ \midrule
1                     & 0.53            & 0.48            & 0.79            & $0.60_{0.13}$    \\
2                     & 0.58            & 0.57            & 0.62            & $0.59_{0.02}$    \\
3                     & 0.47            & 0.57            & 0.64            & $0.56_{0.07}$    \\
4                     & 0.47            & 0.52            & 0.64            & $0.54_{0.07}$   \\ \bottomrule
\end{tabular}}
\caption{The results of each and average accuracy of the selected top four annotators in the three exams, with standard deviations as subscripts in the last column.}
\label{tab:annotator_exam}
\end{table}

\section{Experiment}

\subsection{Implementation Details}
\label{appendix:implementation}

All experiments in this study were conducted on a server equipped with eight NVIDIA A100 GPUs (80GB each). The training code was implemented using the LLAMA-Factory framework~\citep{zheng2024llamafactory}. Table~\ref{tab:hyperparameter} summarizes the hyper-parameters and training configurations for full-parameter fine-tuning.


\begin{table}[]
\centering
\scalebox{0.76}{
\begin{tabular}{lc}
\toprule
\textbf{Hyper-Parameter}        & \textbf{FPFT/LoRA}  \\ \midrule
per\_device\_train\_batch\_size & 4                 \\
gradient\_accumulation\_steps   & 4                   \\
learning\_rate                  & 5.0e-7            \\
num\_train\_epochs              & 10                   \\
lr\_scheduler\_type             & cosine            \\
warmup\_ratio                   & 0.1                \\
optimizer                           & AdamW           \\ \bottomrule   
\end{tabular}}
\caption{Hyper-parameters and training configurations for fine-tuning.}
\label{tab:hyperparameter}
\end{table}

The prompt used in this study to identify client resistance comprised several essential components: a role definition, a task description, examples (optional), a taxonomy of resistance behavior subtypes, a counseling dialogue excerpt that included both contextual exchanges and the target client utterance, and explicit instructions specifying the expected output format. For few-shot prompting of the baseline models, we employ a one-shot strategy, where one example from each category in the training set is randomly selected and incorporated into the prompt as illustrative demonstrations. Specifically, the prompt templates for binary classification and fine-grained resistance classification are shown as Table~\ref{tab:binary_prompt} and Table~\ref{tab:resistance_prompt} respectively.

\begin{table*}[]
    \centering
    \scalebox{0.8}{
    \begin{tabular}{p{20cm}}
    \toprule

\textbf{Role:}

You are a highly professional psychological counselor. You are able to sensitively distinguish client resistance behaviors from collaboration during counseling sessions.

\textbf{Task:}

You will be provided with a snippet of a psychological counseling dialogue between a counselor and a client, including contextual exchanges and a specific client response, as well as a taxonomy of client resistance and collaboration behaviors.
Please carefully read the context and determine the *single most appropriate behavior category* for the client's response based on the provided taxonomy.

\textbf{Resistance Taxonomy}

1. Resistance:

1.1 Arguing: The client challenges the counselor's professionalism or integrity, questions their qualifications or experience, expresses skepticism about the appropriateness of their interventions, or conveys dissatisfaction by implying that the counselor does not truly understand their situation. 

Arguing - Challenging: The client directly questions the accuracy or truthfulness of the information or content provided by the counselor.

Arguing - Discounting: The client questions the counselor's personal abilities, professional knowledge, or role in the counseling process.

1.2 Denying: The client expresses unwillingness or inability to recognize problems, cooperate, accept responsibility, or take advice.

Denying - Blaming: The client expresses resistance by deflecting responsibility and attributing the issue to others, thus refusing to align with the counselor's approach, particularly when the counselor offers suggestions for change.

Denying - Disagreeing: When the counselor employs directive strategies—such as encouraging reflection, helping the client view the problem from a new perspective, suggesting possible solutions, or offering direct advice—the client demonstrates resistance by expressing disagreement without offering any constructive alternatives.

Denying - Excusing: The client conveys their inability to follow the counselor's guidance by offering excuses for their behavior and attributing the issue to external, objective factors.

Denying - Minimizing: The client implies that the counselor has exaggerated the risks or dangers, downplaying the situation as less severe than suggested, and conveying that their issues are not serious enough to warrant the counselor's guidance or assistance.

Denying - Pessimism: The client expresses pessimistic, defeatist, or negative views about themselves, signaling their inability to follow the counselor's suggested course of action.

Denying - Reluctance: The client shows hesitation or expresses reservations toward the information or suggestions offered by the counselor.

Denying - Unwillingness: The client expresses contentment with the current situation and shows a lack of willingness to change, or conveys a resistance to making any changes.

1.3 Avoiding: The client provides brief responses or withholds replies to minimize communication with the counselor on the current topic, thereby avoiding deeper self-exploration and problem-solving.

Avoiding - Minimal Talk: The client withholds information, providing brief and superficial responses that lack depth and sufficient detail, often conveying a dismissive attitude.

Avoiding - Limit Setting: The client explicitly refuses or sidesteps discussing certain topics, often providing alternative reasons for doing so.

1.4 Ignoring: The client shows signs of neglect or non-compliance with the counselor's guidance. The client's responses often suggest that they have not followed the counselor's approaches. It may appear as though the client has completely disregarded the counselor's words or is acting as if the counselor's statements or questions were never made.

Ignoring - Inattention: The client remains fixated on their previously chosen topic, deeply immersed in the original subject or emotions, disregarding the counselor's attempts to intervene.

Ignoring - Sidetracking: The client shifts the conversation away from the counselor's intended focus, introducing new topics or areas of attention.

2. Cooperation: The client aligns with the counselor's direction and engages in the counseling process without resistance.

\textbf{Examples (Optional):}

\textless examples \textgreater

\textbf{Counseling Dialogue:}

Context: \textless context\textgreater

Client Response: \textless client response\textgreater

\textbf{Output Format}

Please provide two lines in the following format: Line 1 starts with "Behavior:" followed by the predicted category; Line 2 starts with "Reason:" followed by a brief justification for the choice. ([option2 - with explanations])

The behavior must be one of the following:
"Resistance" or "Cooperation". 

\\ 
\bottomrule
    \end{tabular}}
    \caption{Prompt template for model-based binary classification task. The prompt settings vary by the presence of examples and the required output format. Zero-shot prompts include no examples, while few-shot prompts provide task-specific exemplars.}
    \label{tab:binary_prompt}
\end{table*}

\begin{table*}[]
    \centering
    \scalebox{0.8}{
    \begin{tabular}{p{20cm}}
    \toprule

\textbf{Role:}

You are a highly professional psychological counselor. You are able to sensitively detect resistant behaviors exhibited by clients during counseling sessions and accurately categorize these behaviors.

\textbf{Task:}

You will be provided with a snippet of a psychological counseling dialogue between a counselor and a client, including contextual exchanges and a specific client response, as well as a taxonomy of client resistance behaviors.
Please carefully read the context and determine the *single most appropriate behavior category* for the client's response based on the provided taxonomy.

\textbf{Resistance Taxonomy}

1. Arguing: The client challenges the counselor's professionalism or integrity, questions their qualifications or experience, expresses skepticism about the appropriateness of their interventions, or conveys dissatisfaction by implying that the counselor does not truly understand their situation. 

Arguing - Challenging: The client directly questions the accuracy or truthfulness of the information or content provided by the counselor.

Arguing - Discounting: The client questions the counselor's personal abilities, professional knowledge, or role in the counseling process.

2. Denying: The client expresses unwillingness or inability to recognize problems, cooperate, accept responsibility, or take advice.

Denying - Blaming: The client expresses resistance by deflecting responsibility and attributing the issue to others, thus refusing to align with the counselor's approach, particularly when the counselor offers suggestions for change.

Denying - Disagreeing: When the counselor employs directive strategies—such as encouraging reflection, helping the client view the problem from a new perspective, suggesting possible solutions, or offering direct advice—the client demonstrates resistance by expressing disagreement without offering any constructive alternatives.

Denying - Excusing: The client conveys their inability to follow the counselor's guidance by offering excuses for their behavior and attributing the issue to external, objective factors.

Denying - Minimizing: The client implies that the counselor has exaggerated the risks or dangers, downplaying the situation as less severe than suggested, and conveying that their issues are not serious enough to warrant the counselor's guidance or assistance.

Denying - Pessimism: The client expresses pessimistic, defeatist, or negative views about themselves, signaling their inability to follow the counselor's suggested course of action.

Denying - Reluctance: The client shows hesitation or expresses reservations toward the information or suggestions offered by the counselor.

Denying - Unwillingness: The client expresses contentment with the current situation and shows a lack of willingness to change, or conveys a resistance to making any changes.

3. Avoiding: The client provides brief responses or withholds replies to minimize communication with the counselor on the current topic, thereby avoiding deeper self-exploration and problem-solving.

Avoiding - Minimal Talk: The client withholds information, providing brief and superficial responses that lack depth and sufficient detail, often conveying a dismissive attitude.

Avoiding - Limit Setting: The client explicitly refuses or sidesteps discussing certain topics, often providing alternative reasons for doing so.

4. Ignoring: The client shows signs of neglect or non-compliance with the counselor's guidance. The client's responses often suggest that they have not followed the counselor's approaches. It may appear as though the client has completely disregarded the counselor's words or is acting as if the counselor's statements or questions were never made.

Ignoring - Inattention: The client remains fixated on their previously chosen topic, deeply immersed in the original subject or emotions, disregarding the counselor's attempts to intervene.

Ignoring - Sidetracking: The client shifts the conversation away from the counselor's intended focus, introducing new topics or areas of attention.

\textbf{Examples (Optional):}

\textless examples \textgreater

\textbf{Counseling Dialogue:}

Context: \textless context\textgreater

Client Response: \textless client response\textgreater

\textbf{Output Format}

Please provide two lines in the following format: Line 1 starts with "Behavior:" followed by the predicted category; Line 2 starts with "Reason:" followed by a brief justification for the choice.  ([option1 - with explanations])

The behavior must be one of the following:
"Arguing - Challenging", "Arguing - Discounting", "Denying - Blaming", "Denying - Disagreeing", "Denying - Excusing", "Denying - Minimizing", "Denying - Pessimism", "Denying - Reluctance", "Denying - Unwillingness", "Avoidance - Minimum Talk", "Avoidance - Limit Setting", "Ignoring - Inattention", or "Ignoring - Sidetracking". 
\\ 
\bottomrule
    \end{tabular}}
    \caption{Prompt template for model-based fine-grained resistance classification task. The prompt settings vary by the presence of examples and the required output format. Zero-shot prompts include no examples, while few-shot prompts provide task-specific exemplars.}
    \label{tab:resistance_prompt}
\end{table*}

\subsection{Experimental Results}
\label{appendix: results}

Table~\ref{tab:appendix_fine_classification} report the per-category performance of all baseline models and our RECAP on the 13-class fine-grained resistance behavior classification task.

Table~\ref{appendix: ablation_results} show the per-category performance of ablation study results on the effects of rationale inclusion and training strategies on Llama-3.1-8B-Instruct and Qwen2.5-7B-Instruct.


\begin{table*}[]
\centering
\scalebox{0.56}{
\begin{tabular}{llccccccccc}
\toprule
\multirow{2}{*}{}          & \multirow{2}{*}{\textbf{Model Name}} & \multicolumn{3}{c}{\textbf{Arguing-Challenging}}                                  & \multicolumn{3}{c}{\textbf{Arguing-Discounting}}                                  & \multicolumn{3}{c}{\textbf{Denying-Blaming}}                                      \\
                           &                                      & \textbf{Precision}        & \textbf{Recall}           & \textbf{F1}               & \textbf{Precision}        & \textbf{Recall}           & \textbf{F1}               & \textbf{Precision}        & \textbf{Recall}           & \textbf{F1}               \\ \midrule
\multirow{8}{*}{\rotatebox{90}{Zero-Shot}} & GPT-4o                            & $34.30 \pm 9.07$          & $29.91 \pm 8.52$          & $31.76 \pm 8.24$          & $84.28 \pm 7.83$          & $40.20 \pm 5.38$          & $54.25 \pm 5.49$          & $34.16 \pm 5.67$          & $42.12 \pm 13.37$         & $37.37 \pm 8.48$          \\
                           & Claude-3.5                        & $34.21 \pm 4.72$          & $48.38 \pm 6.54$          & $40.05 \pm 5.31$          & \textbf{$87.87 \pm 5.68$} & $36.56 \pm 4.12$          & $51.42 \pm 3.88$          & $51.72 \pm 8.39$          & $35.49 \pm 7.48$          & $41.84 \pm 7.17$          \\
                           & Qwen2.5-7B                        & $26.01 \pm 3.54$          & $33.41 \pm 8.44$          & $28.89 \pm 4.76$          & $47.34 \pm 6.50$          & $24.33 \pm 4.04$          & $31.68 \pm 2.93$          & $35.27 \pm 12.66$         & $23.18 \pm 9.97$          & $25.30 \pm 6.02$          \\
                           & Qwen2.5-14B                       & $24.58 \pm 6.04$          & $16.12 \pm 4.24$          & $19.34 \pm 4.79$          & $75.27 \pm 18.78$         & $7.68 \pm 2.11$           & $13.89 \pm 3.71$          & $51.97 \pm 23.77$         & $20.83 \pm 9.20$          & $28.32 \pm 11.12$         \\
                           & Qwen2.5-32B                       & $19.37 \pm 4.55$          & $30.18 \pm 9.10$          & $23.53 \pm 6.08$          & $76.00 \pm 25.10$         & $2.19 \pm 1.04$           & $4.22 \pm 1.95$           & $34.85 \pm 5.07$          & $48.77 \pm 7.18$          & $40.42 \pm 4.68$          \\
                           & Qwen2.5-72B                       & $28.77 \pm 7.41$          & $16.99 \pm 5.72$          & $21.32 \pm 6.53$          & $81.98 \pm 6.79$          & $29.25 \pm 2.21$          & $43.05 \pm 2.80$          & $52.22 \pm 10.73$         & $25.14 \pm 3.75$          & $33.68 \pm 4.51$          \\
                           & Llama-3.1-8B                      & $23.07 \pm 0.91$          & $43.53 \pm 1.97$          & $30.12 \pm 0.68$          & $43.28 \pm 0.66$          & $19.45 \pm 2.37$          & $26.78 \pm 2.44$          & $30.15 \pm 1.49$          & $30.00 \pm 2.13$          & $30.06 \pm 1.58$          \\
                           & Llama-3.1-70B                     & $38.29 \pm 3.39$          & $34.00 \pm 8.16$          & $35.68 \pm 5.66$          & $68.81 \pm 2.03$          & $54.66 \pm 1.54$          & $60.89 \pm 0.89$          & $35.17 \pm 8.66$          & $49.72 \pm 15.96$         & $39.62 \pm 7.33$          \\ \midrule
\multirow{8}{*}{\rotatebox{90}{Few-Shot}}  & GPT-4o                            & $33.74 \pm 9.39$          & $25.23 \pm 5.47$          & $28.66 \pm 6.39$          & $86.26 \pm 5.67$          & $42.57 \pm 6.78$          & $56.72 \pm 5.99$          & $36.16 \pm 4.66$          & $40.27 \pm 8.62$          & $37.87 \pm 5.90$          \\
                           & Claude-3.5                        & $34.73 \pm 3.40$          & $36.64 \pm 3.51$          & $35.65 \pm 3.43$          & $87.86 \pm 5.26$          & $39.11 \pm 2.61$          & $54.08 \pm 2.95$          & $48.77 \pm 9.13$          & $27.02 \pm 6.89$          & $34.33 \pm 7.43$          \\
                           & Qwen2.5-7B                        & $23.31 \pm 4.51$          & $50.72 \pm 8.75$          & $31.92 \pm 5.89$          & $39.21 \pm 12.41$         & $12.23 \pm 4.31$          & $18.63 \pm 6.39$          & $26.10 \pm 8.01$          & $23.75 \pm 10.04$         & $24.75 \pm 9.08$          \\
                           & Qwen2.5-14B                       & $23.57 \pm 6.67$          & $14.36 \pm 3.48$          & $17.74 \pm 4.36$          & $74.69 \pm 19.46$         & $8.23 \pm 2.97$           & $14.79 \pm 5.16$          & $39.22 \pm 11.40$         & $14.67 \pm 6.32$          & $20.95 \pm 7.84$          \\
                           & Qwen2.5-32B                       & $17.28 \pm 5.52$          & $21.70 \pm 7.06$          & $19.23 \pm 6.19$          & $76.00 \pm 25.10$         & $1.83 \pm 1.29$           & $3.55 \pm 2.46$           & $34.80 \pm 5.35$          & $57.36 \pm 6.55$          & $43.23 \pm 5.62$          \\
                           & Qwen2.5-72B                       & $27.07 \pm 11.15$         & $15.82 \pm 6.45$          & $19.93 \pm 8.05$          & $83.06 \pm 7.37$          & $29.97 \pm 2.51$          & $44.01 \pm 3.39$          & $45.12 \pm 15.94$         & $17.56 \pm 7.32$          & $25.18 \pm 9.89$          \\
                           & Llama-3.1-8B                      & $26.28 \pm 4.60$          & $54.96 \pm 7.54$          & $35.54 \pm 5.75$          & $44.98 \pm 5.91$          & $28.98 \pm 5.28$          & $35.23 \pm 5.68$          & $34.80 \pm 15.19$         & $12.44 \pm 3.63$          & $18.16 \pm 5.58$          \\
                           & Llama-3.1-70B                     & $39.26 \pm 2.77$          & $38.12 \pm 6.55$          & $38.54 \pm 4.47$          & $69.99 \pm 2.11$          & $53.74 \pm 2.80$          & $60.78 \pm 2.40$          & $33.66 \pm 7.04$          & $51.67 \pm 7.51$          & $40.72 \pm 7.47$          \\ \midrule
                           & \textbf{Our Model}                   & \bm{$54.60 \pm 4.60$} & \bm{$48.99 \pm 7.48$} & \bm{$51.31 \pm 4.20$} & $76.05 \pm 4.58$          & \bm{$77.86 \pm 9.80$} & \bm{$76.46 \pm 4.31$} & \bm{$65.48 \pm 4.70$} & \bm{$67.28 \pm 4.10$} & \bm{$66.20 \pm 2.39$} \\ \bottomrule
\end{tabular}}
\end{table*}

\begin{table*}[]
\centering
\ContinuedFloat
\scalebox{0.55}{
\begin{tabular}{llccccccccc}
\toprule
                            &                                       & \multicolumn{3}{c}{\textbf{Denying-Disagreeing}}                                  & \multicolumn{3}{c}{\textbf{Denying-Excusing}}                                     & \multicolumn{3}{c}{\textbf{Denying-Minimizing}}                                   \\
\multirow{-2}{*}{}          & \multirow{-2}{*}{\textbf{Model Name}} & \textbf{Precision}        & \textbf{Recall}           & \textbf{F1}               & \textbf{Precision}        & \textbf{Recall}           & \textbf{F1}               & \textbf{Precision}        & \textbf{Recall}           & \textbf{F1}               \\ \midrule
                            & GPT-4o                                & $52.24 \pm 2.85$          & $39.81 \pm 3.67$          & $45.03 \pm 2.12$          & $57.69 \pm 2.55$          & $40.45 \pm 5.09$          & $47.40 \pm 3.64$          & $46.10 \pm 6.29$          & \bm{$77.73 \pm 7.92$} & $57.78 \pm 6.66$          \\
                            & Claude-3.5                            & $57.67 \pm 4.77$          & $36.51 \pm 4.21$          & $44.60 \pm 4.04$          & $30.46 \pm 5.63$ & $52.96 \pm 13.20$         & $38.57 \pm 7.84$          & $53.71 \pm 12.12$         & $46.73 \pm 13.51$         & $47.79 \pm 3.97$          \\
                            & Qwen2.5-7B                            & $27.60 \pm 6.75$          & $45.60 \pm 21.46$         & $30.77 \pm 3.74$          & $14.47 \pm 8.93$          & $16.01 \pm 12.19$         & $14.98 \pm 10.19$         & $24.53 \pm 16.37$         & $34.19 \pm 33.85$         & $20.74 \pm 13.05$         \\
                            & Qwen2.5-14B                           & $27.15 \pm 1.33$          & $47.82 \pm 16.49$         & $33.77 \pm 6.14$          & $14.23 \pm 1.93$          & $59.47 \pm 12.48$         & $22.65 \pm 2.08$          & $18.22 \pm 6.47$          & $71.74 \pm 13.58$         & $28.31 \pm 8.68$          \\
                            & Qwen2.5-32B                           & $47.77 \pm 4.46$          & $25.55 \pm 10.78$         & $32.14 \pm 9.77$          & $21.69 \pm 3.49$          & $64.78 \pm 15.51$         & $32.16 \pm 4.82$          & $30.32 \pm 6.36$          & $51.04 \pm 14.79$         & $36.54 \pm 4.05$          \\
                            & Qwen2.5-72B                           & $46.65 \pm 4.49$          & $20.37 \pm 6.88$          & $28.04 \pm 7.32$          & $42.71 \pm 10.00$         & $36.26 \pm 7.32$          & $38.11 \pm 3.24$          & $33.29 \pm 9.19$          & $63.08 \pm 8.30$          & $42.63 \pm 8.61$          \\
                            & Llama-3.1-8B  & $26.99 \pm 0.56$          & $54.47 \pm 0.98$          & $36.09 \pm 0.61$          & $28.96 \pm 4.18$          & $32.73 \pm 2.54$          & $30.70 \pm 3.48$          & $13.85 \pm 1.90$          & $25.95 \pm 2.42$          & $17.97 \pm 1.82$          \\ \midrule
\multirow{-8}{*}{\rotatebox{90}{Zero-Shot}} & Llama-3.1-70B                         & $26.87 \pm 3.28$          & $45.68 \pm 14.38$         & $33.42 \pm 6.06$          & $10.92 \pm 0.79$          & $80.30 \pm 7.30$ & $19.20 \pm 1.27$          & $33.86 \pm 12.06$         & $51.05 \pm 18.94$         & $37.58 \pm 8.17$          \\
                            & GPT-4o                                & $53.40 \pm 3.57$          & $36.16 \pm 1.76$          & $43.09 \pm 2.08$          & \bm{$59.78 \pm 7.20$} & $42.25 \pm 5.48$          & $49.43 \pm 5.87$          & $46.15 \pm 3.23$          & $73.41 \pm 10.27$         & $56.56 \pm 5.15$          \\
                            & Claude-3.5                            & $62.32 \pm 5.44$          & $37.81 \pm 1.27$          & $46.95 \pm 1.19$          & $35.13 \pm 5.68$          & $49.96 \pm 8.40$          & $41.23 \pm 6.73$          & $70.82 \pm 8.48$          & $47.28 \pm 7.28$          & $56.59 \pm 7.39$          \\
                            & Qwen2.5-7B                            & $26.68 \pm 1.39$          & $56.06 \pm 4.20$          & $36.15 \pm 2.09$          & $21.27 \pm 4.01$          & $23.19 \pm 3.65$          & $22.16 \pm 3.80$          & $16.08 \pm 3.85$          & $37.54 \pm 11.38$         & $22.48 \pm 5.77$          \\
                            & Qwen2.5-14B                           & $30.07 \pm 1.84$          & $53.47 \pm 3.91$          & $38.49 \pm 2.46$          & $14.34 \pm 0.66$          & $70.87 \pm 6.83$          & $23.84 \pm 1.23$          & $17.85 \pm 1.85$          & $67.45 \pm 9.39$          & $28.19 \pm 2.94$          \\
                            & Qwen2.5-32B                           & $46.10 \pm 7.14$          & $25.55 \pm 4.41$          & $32.86 \pm 5.44$          & $27.72 \pm 3.81$          & $59.47 \pm 11.68$         & $37.76 \pm 5.71$          & $32.22 \pm 2.57$          & $52.67 \pm 7.44$          & $39.94 \pm 4.04$          \\
                            & Qwen2.5-72B                           & $46.97 \pm 4.78$          & $19.08 \pm 2.42$          & $27.04 \pm 2.62$          & $41.36 \pm 5.16$          & $39.25 \pm 7.95$          & $39.91 \pm 5.09$          & $35.55 \pm 2.69$          & $64.68 \pm 6.54$          & $45.84 \pm 3.58$          \\
                            & Llama-3.1-8B  & $27.26 \pm 2.64$          & $53.60 \pm 4.30$          & $36.10 \pm 2.97$          & $26.43 \pm 2.56$          & $42.13 \pm 13.40$         & $32.14 \pm 5.55$          & $43.58 \pm 10.30$         & $21.10 \pm 6.78$          & $28.05 \pm 7.09$          \\
\multirow{-8}{*}{\rotatebox{90}{Few-Shot}}  & Llama-3.1-70B                         & $32.53 \pm 2.34$          & $58.18 \pm 4.56$          & $41.73 \pm 3.08$          & $16.82 \pm 0.89$          & $76.79 \pm 3.78$          & $27.57 \pm 1.13$          & $40.03 \pm 2.68$          & $64.16 \pm 7.59$          & $49.19 \pm 3.69$          \\ \midrule
                            & \textbf{Our Model}                    & \bm{$69.19 \pm 4.41$} & \bm{$69.85 \pm 6.22$} & \bm{$69.38 \pm 4.13$} & $58.53 \pm 8.41$          & \bm{$58.34 \pm 6.92$} & \bm{$58.26 \pm 6.83$} & \bm{$70.22 \pm 6.58$} & $73.38 \pm 8.38$ & \bm{$71.74 \pm 7.37$} \\ \bottomrule
\end{tabular}}
\end{table*}

\begin{table*}[]
\centering
\ContinuedFloat
\scalebox{0.56}{
\begin{tabular}{llccccccccc}
\toprule
\multirow{2}{*}{}          & \multirow{2}{*}{\textbf{Model Name}} & \multicolumn{3}{c}{\textbf{Denying-Pessimism}}                                    & \multicolumn{3}{c}{\textbf{Denying-Relunctance}}                                  & \multicolumn{3}{c}{\textbf{Denying-Unwillingness}}                                 \\
                           &                                      & \textbf{Precision}        & \textbf{Recall}           & \textbf{F1}               & \textbf{Precision}        & \textbf{Recall}           & \textbf{F1}               & \textbf{Precision}         & \textbf{Recall}           & \textbf{F1}               \\ \midrule
\multirow{8}{*}{\rotatebox{90}{Zero-Shot}} & GPT-4o                               & $44.26 \pm 2.79$          & $74.32 \pm 4.38$          & $55.36 \pm 1.92$          & $40.35 \pm 3.65$          & $57.00 \pm 5.31$          & $47.07 \pm 2.96$          & $25.08 \pm 1.61$           & $64.62 \pm 4.00$ & $36.09 \pm 1.83$          \\
                           & Claude-3.5                           & $44.83 \pm 3.78$          & $77.12 \pm 4.93$          & $56.55 \pm 2.78$          & $47.66 \pm 2.62$ & $30.75 \pm 6.35$          & $36.99 \pm 4.57$          & $49.08 \pm 5.10$           & $27.07 \pm 3.86$          & $34.68 \pm 3.60$          \\
                           & Qwen2.5-7B                           & $47.11 \pm 7.02$          & $20.80 \pm 8.46$          & $27.72 \pm 7.86$          & $39.86 \pm 12.08$         & $17.74 \pm 13.75$         & $20.10 \pm 12.75$         & $21.64 \pm 7.79$           & $33.85 \pm 19.48$         & $23.16 \pm 5.78$          \\
                           & Qwen2.5-14B                          & $47.46 \pm 1.75$          & $46.96 \pm 4.80$          & $47.09 \pm 2.64$          & $24.93 \pm 1.42$          & $59.34 \pm 8.05$          & $34.95 \pm 1.58$          & $36.55 \pm 5.58$           & $26.07 \pm 10.37$         & $29.46 \pm 7.79$          \\
                           & Qwen2.5-32B                          & $32.14 \pm 2.20$          & $81.33 \pm 4.85$          & $46.02 \pm 2.45$          & $27.73 \pm 3.72$          & $56.60 \pm 8.96$          & $36.87 \pm 3.60$          & $23.98 \pm 3.09$           & $53.13 \pm 4.76$          & $33.00 \pm 3.59$          \\
                           & Qwen2.5-72B                          & $43.93 \pm 2.62$          & $71.74 \pm 6.16$          & $54.34 \pm 2.38$          & $41.52 \pm 2.41$          & $46.45 \pm 4.48$          & $43.77 \pm 2.69$          & $38.05 \pm 9.48$           & $31.75 \pm 6.64$          & $34.33 \pm 7.23$          \\
                           & Llama-3.1-8B                         & $41.84 \pm 0.99$          & $31.16 \pm 2.24$          & $35.70 \pm 1.81$          & $29.95 \pm 2.06$          & $33.22 \pm 0.93$          & $31.49 \pm 1.55$          & $21.69 \pm 0.66$           & $48.95 \pm 1.44$          & $30.05 \pm 0.76$          \\
                           & Llama-3.1-70B                        & $58.51 \pm 9.03$          & $26.64 \pm 3.03$          & $36.24 \pm 2.12$          & $69.43 \pm 12.62$         & $12.62 \pm 6.85$ & $20.99 \pm 10.26$         & $51.27 \pm 12.02$          & $13.52 \pm 10.00$         & $20.01 \pm 12.50$         \\ \midrule
\multirow{8}{*}{\rotatebox{90}{Few-Shot}}  & GPT-4o                               & $42.70 \pm 2.16$          & \bm{$77.58 \pm 1.40$}          & $55.06 \pm 1.98$          & $42.08 \pm 2.95$ & $55.61 \pm 5.84$          & $47.86 \pm 3.84$          & $24.35 \pm 1.45$           & \bm{$65.60 \pm 4.08$} & $35.48 \pm 1.68$          \\
                           & Claude-3.5                           & $44.39 \pm 3.44$          & $68.94 \pm 6.61$          & $53.99 \pm 4.55$          & $46.09 \pm 2.73$          & $30.02 \pm 3.18$          & $36.30 \pm 2.71$          & $35.87 \pm 5.74$           & $34.90 \pm 6.06$          & $35.34 \pm 5.79$          \\
                           & Qwen2.5-7B                           & $42.80 \pm 2.90$          & $33.42 \pm 3.42$          & $37.44 \pm 2.58$          & $26.38 \pm 3.09$          & $33.76 \pm 5.10$          & $29.58 \pm 3.72$          & $20.80 \pm 2.34$           & $46.90 \pm 6.57$          & $28.78 \pm 3.32$          \\
                           & Qwen2.5-14B                          & $43.46 \pm 2.26$          & $53.29 \pm 6.92$          & $47.77 \pm 3.72$          & $26.25 \pm 3.29$          & $53.95 \pm 6.16$          & $35.31 \pm 4.28$          & $41.00 \pm 14.13$          & $21.32 \pm 6.39$          & $27.88 \pm 8.32$          \\
                           & Qwen2.5-32B                          & $34.51 \pm 1.08$          & $77.34 \pm 4.25$ & $47.71 \pm 1.73$          & $30.11 \pm 2.80$          & $61.40 \pm 5.41$          & $40.39 \pm 3.65$          & $16.87 \pm 1.51$           & $60.43 \pm 4.80$          & $26.36 \pm 2.18$          \\
                           & Qwen2.5-72B                          & $46.23 \pm 1.79$          & $72.67 \pm 2.21$          & $56.51 \pm 1.95$          & $40.41 \pm 1.82$          & $55.27 \pm 7.21$          & $46.58 \pm 3.67$          & $33.81 \pm 8.89$           & $36.49 \pm 8.43$          & $35.05 \pm 8.58$          \\
                           & Llama-3.1-8B                         & $40.99 \pm 6.46$          & $43.01 \pm 7.56$          & $41.92 \pm 6.85$          & $58.04 \pm 18.01$         & $7.69 \pm 2.17$           & $13.29 \pm 3.35$          & \bm{$62.64 \pm 6.81$}  & $16.23 \pm 2.44$          & $25.72 \pm 3.42$          \\
                           & Llama-3.1-70B                        & $63.14 \pm 5.11$          & $25.01 \pm 5.62$          & $35.60 \pm 6.46$          & $63.85 \pm 10.36$         & $22.51 \pm 7.67$          & $33.08 \pm 9.79$          & $36.47 \pm 6.02$           & $22.42 \pm 5.48$          & $27.69 \pm 5.92$          \\ \midrule
                           & \textbf{Our Model}                   & \bm{$65.84 \pm 5.30$} & $69.15 \pm 6.59$ & \bm{$67.24 \pm 4.03$} & \bm{$74.00 \pm 4.76$} & \bm{$68.93 \pm 6.55$} & \bm{$71.30 \pm 5.32$} & \bm{$60.48 \pm 10.51$} & $63.45 \pm 9.25$ & \bm{$61.35 \pm 6.73$} \\ \bottomrule
\end{tabular}}
\end{table*}

\begin{table*}[]
\centering
\scalebox{0.55}{
\begin{tabular}{llccccccccc}
\toprule
\multirow{2}{*}{}          & \multirow{2}{*}{\textbf{Model Name}} & \multicolumn{3}{c}{\textbf{Avoidance-Minimum Talk}}                           & \multicolumn{3}{c}{\textbf{Avoidance-Limit Setting}}                              & \multicolumn{3}{c}{\textbf{Ignoring-Sidetracking}}                                  \\
                           &                                      & \textbf{Precision}        & \textbf{Recall}           & \textbf{F1}               & \textbf{Precision}        & \textbf{Recall}           & \textbf{F1}               & \textbf{Precision}        & \textbf{Recall}           & \textbf{F1}               \\ \midrule
\multirow{8}{*}{\rotatebox{90}{Zero-Shot}} & GPT-4o                               & $51.37 \pm 2.70$          & $79.17 \pm 3.38$          & $62.26 \pm 2.27$          & $55.52 \pm 8.48$          & $32.00 \pm 5.16$          & $40.50 \pm 6.08$          & $61.82 \pm 3.65$          & $62.95 \pm 5.15$ & $62.22 \pm 2.59$          \\
                           & Claude-3.5                           & $44.99 \pm 2.11$          & \bm{$87.21 \pm 6.16$} & $59.24 \pm 1.64$          & $39.08 \pm 3.99$ & $46.82 \pm 13.29$         & $41.81 \pm 5.67$          & $52.85 \pm 11.01$         & $61.85 \pm 6.13$          & $56.09 \pm 3.77$          \\
                           & Qwen2.5-7B                           & $29.44 \pm 41.18$         & $0.98 \pm 0.98$           & $1.87 \pm 1.86$           & $64.78 \pm 22.35$         & $13.65 \pm 12.55$         & $18.85 \pm 14.00$         & $48.68 \pm 22.48$         & $35.57 \pm 28.29$         & $30.28 \pm 5.06$          \\
                           & Qwen2.5-14B                          & $47.38 \pm 15.16$         & $6.89 \pm 5.54$           & $11.32 \pm 8.75$          & $59.39 \pm 25.77$         & $12.24 \pm 5.56$          & $18.95 \pm 6.29$          & $64.74 \pm 10.57$         & $28.20 \pm 16.06$         & $35.92 \pm 12.04$         \\
                           & Qwen2.5-32B                          & $42.14 \pm 3.83$          & $38.12 \pm 8.43$          & $39.60 \pm 4.87$          & $46.69 \pm 4.73$          & $26.82 \pm 7.60$          & $33.42 \pm 5.15$          & $77.63 \pm 4.97$          & $28.18 \pm 3.96$          & $41.18 \pm 4.29$          \\
                           & Qwen2.5-72B                          & $35.73 \pm 3.10$          & $80.74 \pm 6.97$          & $49.30 \pm 1.83$          & $27.27 \pm 6.56$          & $38.12 \pm 7.28$          & $31.15 \pm 5.11$          & $44.64 \pm 2.62$          & $66.36 \pm 4.99$          & $53.28 \pm 2.39$          \\
                           & Llama-3.1-8B                         & $0.00 \pm 0.00$           & $0.00 \pm 0.00$           & $0.00 \pm 0.00$           & $59.40 \pm 1.70$          & $11.76 \pm 1.44$          & $19.62 \pm 2.12$          & $55.41 \pm 3.27$          & $42.45 \pm 4.52$          & $47.82 \pm 2.01$          \\
                           & Llama-3.1-70B                        & $51.33 \pm 4.69$          & $37.53 \pm 15.15$         & $42.13 \pm 11.88$         & \bm{$80.97 \pm 14.82$}         & $17.18 \pm 6.69$ & $27.45 \pm 7.74$          & $72.49 \pm 3.06$          & $40.03 \pm 7.50$          & $51.10 \pm 5.41$          \\ \midrule
\multirow{8}{*}{\rotatebox{90}{Few-Shot}}  & GPT-4o                               & $49.78 \pm 2.39$          & $84.87 \pm 2.29$          & $62.73 \pm 2.21$          & $57.12 \pm 6.90$ & $32.00 \pm 5.79$          & $40.98 \pm 6.42$          & $60.77 \pm 1.98$          & $64.09 \pm 3.28$ & $62.37 \pm 2.47$          \\
                           & Claude-3.5                           & $45.50 \pm 2.43$          & $92.33 \pm 3.73$          & $60.94 \pm 2.76$          & $49.65 \pm 1.87$          & $41.88 \pm 4.45$          & $45.38 \pm 3.20$          & $46.71 \pm 2.80$          & $73.12 \pm 2.58$          & $56.99 \pm 2.71$          \\
                           & Qwen2.5-7B                           & $9.17 \pm 14.55$          & $0.39 \pm 0.54$           & $0.74 \pm 1.02$           & $61.94 \pm 8.12$          & $9.41 \pm 0.83$           & $16.33 \pm 1.50$          & $56.93 \pm 5.87$          & $34.38 \pm 7.13$          & $42.72 \pm 6.71$          \\
                           & Qwen2.5-14B                          & $50.78 \pm 22.64$         & $4.32 \pm 2.03$           & $7.89 \pm 3.58$           & $65.07 \pm 18.65$         & $10.12 \pm 3.29$          & $17.44 \pm 5.50$          & $62.68 \pm 1.86$          & $50.35 \pm 6.40$          & $55.66 \pm 3.75$          \\
                           & Qwen2.5-32B                          & $46.85 \pm 6.77$          & $46.56 \pm 6.67$ & $46.70 \pm 6.71$          & $54.49 \pm 4.88$          & $22.59 \pm 3.47$          & $31.79 \pm 3.82$          & $73.14 \pm 4.52$          & $44.53 \pm 5.29$          & $55.25 \pm 4.75$          \\
                           & Qwen2.5-72B                          & $37.94 \pm 2.13$          & $81.92 \pm 3.61$          & $51.85 \pm 2.57$          & $33.92 \pm 6.34$          & $32.00 \pm 6.08$          & $32.89 \pm 6.11$          & $36.71 \pm 2.30$          & \bm{$73.49 \pm 2.84$} & $48.94 \pm 2.49$          \\
                           & Llama-3.1-8B                         & $23.21 \pm 8.67$          & $13.00 \pm 6.04$          & $16.58 \pm 7.24$          & $66.61 \pm 12.66$         & $11.18 \pm 4.35$          & $18.87 \pm 6.69$          & $60.72 \pm 7.78$ & $24.65 \pm 4.96$          & $34.92 \pm 5.77$          \\
                           & Llama-3.1-70B                        & $54.91 \pm 3.92$          & $50.09 \pm 6.34$          & $52.30 \pm 4.85$          & $75.01 \pm 5.50$          & $21.41 \pm 3.16$          & $33.17 \pm 3.70$          & $65.83 \pm 4.44$          & $48.86 \pm 3.89$          & $55.93 \pm 2.71$          \\ \midrule
                           & \textbf{Our Model}                   & \bm{$77.48 \pm 3.30$} & $80.73 \pm 5.84$ & \bm{$78.90 \pm 2.31$} & $70.68 \pm 3.92$ & \bm{$68.94 \pm 2.44$} & \bm{$69.70 \pm 1.40$} & \bm{$73.40 \pm 3.66$} & $59.76 \pm 4.03$ & \bm{$65.72 \pm 1.75$} \\ \bottomrule
\end{tabular}}
\end{table*}

\begin{table*}[]
\centering
\scalebox{0.55}{
\begin{tabular}{clccccccc}
\toprule
\multirow{2}{*}{}          & \multirow{2}{*}{\textbf{Model Name}} & \multicolumn{3}{c}{\textbf{Ignoring-Inattention}}                                  & \multicolumn{4}{c}{\textbf{Fine-grained Overall}}                                                             \\
                           &                                      & \textbf{Precision}        & \textbf{Recall}            & \textbf{F1}               & \textbf{Precision}        & \textbf{Recall}           & \textbf{F1}               & \textbf{Acc}              \\ \midrule
\multirow{8}{*}{\rotatebox{90}{Zero-Shot}} & GPT-4o                               & $36.96 \pm 9.63$          & $6.35 \pm 2.21$            & $10.76 \pm 3.47$          & $48.01 \pm 0.98$          & $49.74 \pm 1.23$          & $45.22 \pm 0.63$          & $47.62 \pm 0.50$          \\
                           & Claude-3.5                           & $51.87 \pm 11.12$         & $13.08 \pm 13.83$ & $17.86 \pm 13.26$         & $49.69 \pm 1.17$ & $46.19 \pm 1.03$          & $43.65 \pm 1.52$          & $46.87 \pm 1.22$          \\
                           & Qwen2.5-7B                           & $7.33 \pm 10.31$          & $0.77 \pm 0.80$            & $1.37 \pm 1.48$           & $33.39 \pm 1.91$          & $23.08 \pm 5.60$          & $21.21 \pm 5.60$          & $23.76 \pm 5.43$          \\
                           & Qwen2.5-14B                          & $22.57 \pm 23.40$         & $0.96 \pm 1.18$            & $1.83 \pm 2.21$           & $39.57 \pm 2.70$          & $31.10 \pm 1.99$          & $25.06 \pm 1.97$          & $27.81 \pm 2.46$          \\
                           & Qwen2.5-32B                          & $9.00 \pm 12.45$          & $0.38 \pm 0.53$            & $0.74 \pm 1.01$           & $37.64 \pm 1.85$          & $39.01 \pm 2.48$          & $30.76 \pm 1.33$          & $32.97 \pm 1.31$          \\
                           & Qwen2.5-72B                          & $9.00 \pm 12.45$          & $0.38 \pm 0.53$            & $0.74 \pm 1.01$           & $40.44 \pm 1.41$          & $40.51 \pm 1.22$          & $36.44 \pm 0.77$          & $39.49 \pm 0.94$          \\
                           & Llama-3.1-8B                         & $5.98 \pm 1.05$           & $3.08 \pm 1.05$            & $4.03 \pm 1.16$           & $29.27 \pm 0.81$          & $28.98 \pm 1.02$          & $26.19 \pm 1.30$          & $28.83 \pm 1.09$          \\
                           & Llama-3.1-70B                        & $21.07 \pm 17.04$         & $1.92 \pm 1.52$            & $3.49 \pm 2.75$           & $47.61 \pm 2.05$          & $35.76 \pm 3.11$ & $32.91 \pm 3.30$          & $34.64 \pm 3.53$          \\ \midrule
\multirow{8}{*}{\rotatebox{90}{Few-Shot}}  & GPT-4o                               & $44.71 \pm 6.71$          & $7.69 \pm 1.36$            & $13.02 \pm 1.89$          & $49.00 \pm 1.39$ & $49.79 \pm 0.86$          & $45.37 \pm 0.82$          & $47.84 \pm 0.94$          \\
                           & Claude-3.5                           & $35.09 \pm 8.72$          & $22.12 \pm 5.97$           & $27.11 \pm 7.07$          & $49.45 \pm 1.12$          & $46.24 \pm 1.72$          & $44.99 \pm 1.43$          & $47.89 \pm 1.39$          \\
                           & Qwen2.5-7B                           & $7.90 \pm 4.66$           & $2.31 \pm 1.46$            & $3.56 \pm 2.22$           & $29.12 \pm 1.92$          & $28.01 \pm 1.36$          & $24.25 \pm 0.89$          & $27.75 \pm 1.42$          \\
                           & Qwen2.5-14B                          & $6.04 \pm 5.56$           & $0.58 \pm 0.53$            & $1.05 \pm 0.96$           & $38.08 \pm 2.40$          & $32.54 \pm 1.37$          & $25.92 \pm 1.02$          & $30.49 \pm 0.81$          \\
                           & Qwen2.5-32B                          & $10.15 \pm 10.84$         & $0.58 \pm 0.53$   & $1.08 \pm 0.99$           & $38.48 \pm 2.97$          & $40.92 \pm 2.37$          & $32.76 \pm 1.69$          & $34.99 \pm 1.72$          \\
                           & Qwen2.5-72B                          & $2.50 \pm 5.59$           & $0.19 \pm 0.43$            & $0.36 \pm 0.80$           & $39.28 \pm 1.11$          & $41.42 \pm 1.26$          & $36.47 \pm 0.96$          & $40.20 \pm 0.47$          \\
                           & Llama-3.1-8B                         & $12.89 \pm 2.72$          & $22.36 \pm 4.40$           & $16.28 \pm 3.03$          & $40.65 \pm 1.90$          & $27.02 \pm 1.97$          & $27.14 \pm 2.01$          & $29.54 \pm 2.28$ \\
                           & Llama-3.1-70B                        & $28.49 \pm 5.69$          & $9.23 \pm 1.99$            & $13.91 \pm 2.86$          & $47.69 \pm 1.16$          & $41.71 \pm 1.61$          & $39.25 \pm 1.84$          & $41.24 \pm 1.78$          \\ \midrule
                           & \textbf{Our Model}                   & \bm{$55.31 \pm 2.85$} & \bm{$61.15 \pm 7.77$}  & \bm{$57.93 \pm 4.23$} & \bm{$67.02 \pm 1.37$} & \bm{$66.76 \pm 1.41$} & \bm{$66.58 \pm 1.30$} & \bm{$67.72 \pm 0.99$} \\ \bottomrule
\end{tabular}}
\caption{Fine-Grained Resistance Subtype Classification. Per-category performance metrics (precision, recall, F1) along with overall accuracy and macro-averaged scores.}
\label{tab:appendix_fine_classification}
\end{table*}


\begin{table*}[]
\centering
\scalebox{0.60}{
\begin{tabular}{cccccccccccc}
\toprule
\multirow{2}{*}{\textbf{Backbone}}     & \multirow{2}{*}{\textbf{Rationale}} & \multirow{2}{*}{\textbf{Fine-tune}} & \multicolumn{3}{c}{\textbf{Arguing-Challenging}}      & \multicolumn{3}{c}{\textbf{Arguing-Discounting}}      & \multicolumn{3}{c}{\textbf{Denying-Blaming}}          \\
                                       &                                     &                                     & \textbf{Precision} & \textbf{Recall} & \textbf{F1}    & \textbf{Precision} & \textbf{Recall} & \textbf{F1}    & \textbf{Precision} & \textbf{Recall} & \textbf{F1}    \\ \midrule
\multirow{4}{*}{Llama-3.1-8B-Instruct} &   \ding{51}      & FPFT                                & $54.60_{4.60}$     & $48.99_{7.48}$  & $51.31_{4.20}$ & $76.05_{4.58}$     & $77.86_{9.80}$  & $76.46_{4.31}$ & $65.48_{4.70}$     & $67.28_{4.10}$  & $66.20_{2.39}$ \\
                                       &   \ding{55}         & FPFT                                & $49.07_{7.42}$     & $44.87_{6.94}$  & $46.78_{6.87}$ & $73.88_{5.84}$     & $76.43_{5.50}$  & $74.94_{3.85}$ & $60.44_{3.33}$     & $60.23_{6.17}$  & $60.27_{4.43}$ \\
                                       & \ding{51}         & LoRA                                & $34.48_{1.53}$     & $67.40_{3.89}$  & $45.60_{1.92}$ & $71.54_{4.87}$     & $43.83_{2.82}$  & $54.30_{3.02}$ & $45.16_{10.24}$    & $33.08_{8.56}$  & $38.04_{8.80}$ \\
                                       &      \ding{55}                                & \ding{55}                                  & $23.07_{0.91}$     & $43.53_{1.97}$  & $30.12_{0.68}$ & $43.28_{0.66}$     & $19.45_{2.37}$  & $26.78_{2.44}$ & $30.15_{1.49}$     & $30.00_{2.13}$  & $30.06_{1.58}$ \\ \midrule
\multirow{4}{*}{Qwen2.5-7B-Instruct}   &    \ding{51}                                  & FPFT                                & $50.31_{6.37}$     & $44.59_{7.71}$  & $47.10_{6.52}$ & $74.81_{6.03}$     & $77.52_{3.90}$  & $75.99_{3.54}$ & $62.93_{3.27}$     & $61.16_{4.62}$  & $62.00_{3.69}$ \\
                                       &   \ding{55}                                   & FPFT                                & $42.62_{6.12}$     & $42.26_{13.22}$ & $41.58_{8.10}$ & $69.60_{6.27}$     & $78.61_{5.46}$  & $73.52_{2.66}$ & $60.72_{11.34}$    & $54.03_{4.88}$  & $56.62_{5.30}$ \\
                                       &    \ding{51}                                  & LoRA                                & $31.44_{2.61}$     & $63.37_{6.69}$  & $41.98_{3.46}$ & $65.27_{7.98}$     & $37.43_{3.40}$  & $47.56_{4.85}$ & $35.29_{5.10}$     & $23.08_{4.49}$  & $27.87_{4.84}$ \\
                                       &   \ding{55}                                   & \ding{55}                                   & $26.01_{3.54}$     & $33.41_{8.44}$  & $28.89_{4.76}$ & $47.34_{6.50}$     & $24.33_{4.04}$  & $31.68_{2.93}$ & $35.27_{12.66}$    & $23.18_{9.97}$  & $25.30_{6.02}$ \\ \toprule
\end{tabular}}
\end{table*}

\begin{table*}[]
\centering
\scalebox{0.6}{
\begin{tabular}{cccccccccccc}
\toprule
\multirow{2}{*}{\textbf{Backbone}}     & \multirow{2}{*}{\textbf{Rationale}} & \multirow{2}{*}{\textbf{Fine-tune}} & \multicolumn{3}{c}{\textbf{Denying-Pessimism}}        & \multicolumn{3}{c}{\textbf{Denying-Relunctance}}       & \multicolumn{3}{c}{\textbf{Denying-Unwillingness}}    \\
                                       &                                     &                                     & \textbf{Precision} & \textbf{Recall} & \textbf{F1}    & \textbf{Precision} & \textbf{Recall} & \textbf{F1}     & \textbf{Precision} & \textbf{Recall} & \textbf{F1}    \\ \midrule
\multirow{4}{*}{Llama-3.1-8B-Instruct} &   \ding{51}                                   & FPFT                                & $65.84_{5.30}$     & $69.15_{6.59}$  & $67.24_{4.03}$ & $74.00_{4.76}$     & $68.93_{6.55}$  & $71.30_{5.32}$  & $60.48_{10.51}$    & $63.45_{9.25}$  & $61.35_{6.73}$ \\
                                       &     \ding{55}                                 & FPFT                                & $60.56_{5.16}$     & $63.08_{3.22}$  & $61.71_{3.60}$ & $62.00_{9.18}$     & $63.46_{8.06}$  & $62.59_{8.22}$  & $59.04_{8.19}$     & $48.37_{7.58}$  & $52.71_{5.90}$ \\
                                       &   \ding{51}                                   & LoRA                                & $51.83_{3.71}$     & $62.59_{5.55}$  & $56.60_{3.67}$ & $75.40_{8.69}$     & $31.60_{4.26}$  & $44.14_{3.28}$  & $60.23_{6.96}$     & $40.89_{1.94}$  & $48.53_{2.02}$ \\
                                       &   \ding{55}                                   & \ding{55}                                 & $41.84_{0.99}$     & $31.16_{2.24}$  & $35.70_{1.81}$ & $29.95_{2.06}$     & $33.22_{0.93}$  & $31.49_{1.55}$  & $21.69_{0.66}$     & $48.95_{1.44}$  & $30.05_{0.76}$ \\ \midrule
\multirow{4}{*}{Qwen2.5-7B-Instruct}   &   \ding{51}                                   & FPFT                                & $61.37_{3.44}$     & $64.95_{6.22}$  & $62.95_{3.41}$ & $61.74_{5.27}$     & $66.20_{9.05}$  & $63.75_{6.55}$  & $63.79_{5.71}$     & $52.00_{8.90}$  & $56.83_{5.76}$ \\
                                       & \ding{55}                                     & FPFT                                & $56.58_{5.00}$     & $67.05_{3.67}$  & $61.16_{2.32}$ & $63.99_{9.31}$     & $65.13_{9.90}$  & $63.85_{5.94}$  & $49.31_{10.31}$    & $55.75_{10.55}$ & $51.19_{5.65}$ \\
                                       &   \ding{51}                                   & LoRA                                & $52.25_{4.29}$     & $55.57_{6.98}$  & $53.82_{5.42}$ & $68.67_{8.22}$     & $29.87_{8.22}$  & $40.87_{7.55}$  & $57.63_{7.19}$     & $36.35_{1.78}$  & $44.35_{1.59}$ \\
                                       &  \ding{55}                                    & \ding{55}                                   & $47.11_{7.02}$     & $20.80_{8.46}$  & $27.72_{7.86}$ & $39.86_{12.08}$    & $17.74_{13.75}$ & $20.10_{12.75}$ & $21.64_{7.79}$     & $33.85_{19.48}$ & $23.16_{5.78}$ \\ \bottomrule
\end{tabular}}
\end{table*}

\begin{table*}[]
\centering
\scalebox{0.6}{
\begin{tabular}{cccccccccccc}
\toprule
\multirow{2}{*}{\textbf{Backbone}}     & \multirow{2}{*}{\textbf{Rationale}} & \multirow{2}{*}{\textbf{Fine-tune}} & \multicolumn{3}{c}{\textbf{Avoidance-Minimum Response}} & \multicolumn{3}{c}{\textbf{Avoidance-Limit Setting}}   & \multicolumn{3}{c}{\textbf{Ignoring-Side Track}}      \\
                                       &                                     &                                     & \textbf{Precision}  & \textbf{Recall}  & \textbf{F1}    & \textbf{Precision} & \textbf{Recall} & \textbf{F1}     & \textbf{Precision} & \textbf{Recall} & \textbf{F1}    \\ \midrule
\multirow{4}{*}{Llama-3.1-8B-Instruct} &     \ding{51}                                 & FPFT                                & $77.48_{3.30}$      & $80.73_{5.84}$   & $78.90_{2.31}$ & $70.68_{3.92}$     & $68.94_{2.44}$  & $69.70_{1.40}$  & $73.40_{3.66}$     & $59.76_{4.03}$  & $65.72_{1.75}$ \\
                                       &    \ding{55}                                  & FPFT                                & $73.52_{5.33}$      & $75.43_{4.40}$   & $74.30_{3.10}$ & $68.84_{4.55}$     & $63.29_{2.41}$  & $65.88_{2.50}$  & $68.19_{5.80}$     & $67.49_{6.35}$  & $67.73_{5.38}$ \\
                                       &  \ding{51}                                    & LoRA                                & $56.73_{2.90}$      & $80.08_{4.38}$   & $66.34_{2.42}$ & $73.12_{3.17}$     & $36.47_{5.52}$  & $48.56_{5.74}$  & $64.59_{1.61}$     & $57.26_{4.03}$  & $60.64_{2.27}$ \\
                                       &   \ding{55}                                   & \ding{55}                                   & $0.00_{0.00}$       & $0.00_{0.00}$    & $0.00_{0.00}$  & $59.40_{1.70}$     & $11.76_{1.44}$  & $19.62_{2.12}$  & $55.41_{3.27}$     & $42.45_{4.52}$  & $47.82_{2.01}$ \\ \midrule
\multirow{4}{*}{Qwen2.5-7B-Instruct}   &   \ding{51}                                   & FPFT                                & $74.04_{2.25}$      & $77.20_{4.05}$   & $75.54_{2.46}$ & $71.84_{1.55}$     & $61.18_{4.71}$  & $66.00_{2.87}$  & $69.38_{3.27}$     & $68.80_{6.11}$  & $69.05_{4.58}$ \\
                                       &  \ding{55}                                    & FPFT                                & $71.30_{4.54}$      & $77.59_{7.52}$   & $74.03_{3.22}$ & $70.64_{2.73}$     & $55.53_{4.66}$  & $62.03_{2.55}$  & $70.26_{3.28}$     & $63.17_{8.00}$  & $66.30_{4.91}$ \\
                                       &    \ding{51}                                  & LoRA                                & $55.36_{1.47}$      & $74.67_{5.30}$   & $63.48_{1.57}$ & $72.79_{5.70}$     & $27.06_{4.61}$  & $39.36_{5.64}$  & $68.40_{1.11}$     & $49.76_{2.72}$  & $57.57_{1.86}$ \\
                                       &    \ding{55}                                  & \ding{55}                                   & $29.44_{41.18}$     & $0.98_{0.98}$    & $1.87_{1.86}$  & $64.78_{22.35}$    & $13.65_{12.55}$ & $18.85_{14.00}$ & $48.68_{22.48}$    & $35.57_{28.29}$ & $30.28_{5.06}$ \\ \bottomrule
\end{tabular}}
\end{table*}

\begin{table*}[]
\centering
\scalebox{0.6}{
\begin{tabular}{cccccccccc}
\toprule
\multirow{2}{*}{\textbf{Backbone}}     & \multirow{2}{*}{\textbf{Rationale}} & \multirow{2}{*}{\textbf{Fine-tune}} & \multicolumn{3}{c}{\textbf{Ignoring-Inattention}}     & \multicolumn{4}{c}{\textbf{Fine-grained Overall}}                      \\
                                       &                                     &                                     & \textbf{Precision} & \textbf{Recall} & \textbf{F1}    & \textbf{Precision} & \textbf{Recall} & \textbf{F1}    & \textbf{Acc}   \\ \midrule
\multirow{4}{*}{Llama-3.1-8B-Instruct} &   \ding{51}                                  & FPFT                                & $55.31_{2.85}$     & $61.15_{7.77}$  & $57.93_{4.23}$ & $67.02_{1.37}$     & $66.76_{1.41}$  & $66.58_{1.30}$ & $67.72_{0.99}$ \\
                                       &    \ding{55}                                   & FPFT                                & $51.84_{4.22}$     & $50.38_{8.98}$  & $50.86_{5.98}$ & $62.44_{1.40}$     & $61.42_{1.48}$  & $61.72_{1.52}$ & $63.36_{1.76}$ \\
                                       &    \ding{51}                                   & LoRA                                & $40.93_{10.98}$    & $17.79_{6.50}$  & $24.75_{8.34}$ & $54.81_{1.97}$     & $50.30_{1.62}$  & $49.54_{1.19}$ & $51.42_{1.09}$ \\
                                       &     \ding{55}                                    &  \ding{55}                                     & $5.98_{1.05}$      & $3.08_{1.05}$   & $4.03_{1.16}$  & $29.27_{0.81}$     & $28.98_{1.02}$  & $26.19_{1.30}$ & $28.83_{1.09}$ \\ \midrule
\multirow{4}{*}{Qwen2.5-7B-Instruct}   &    \ding{51}                                 & FPFT                                & $54.19_{2.70}$     & $53.46_{6.58}$  & $53.63_{3.75}$ & $63.90_{1.34}$     & $62.74_{1.50}$  & $62.91_{1.11}$ & $64.40_{1.46}$ \\
                                       &     \ding{55}                                    & FPFT                                & $49.43_{5.32}$     & $53.27_{10.19}$ & $50.87_{5.51}$ & $61.63_{2.80}$     & $59.41_{1.29}$  & $59.49_{1.77}$ & $61.15_{2.22}$ \\
                                       &    \ding{51}                                     & LoRA                                & $35.86_{8.44}$     & $17.31_{4.71}$  & $23.22_{5.91}$ & $51.60_{1.55}$     & $45.34_{1.29}$  & $45.09_{0.91}$ & $46.95_{1.00}$ \\
                                       &   \ding{55}                                      &  \ding{55}                                     & $7.33_{10.31}$     & $0.77_{0.80}$   & $1.37_{1.48}$  & $33.39_{1.91}$     & $23.08_{5.60}$  & $21.21_{5.60}$ & $23.76_{5.43}$ \\ \bottomrule
\end{tabular}}
\caption{Ablation study results on the effects of rationale inclusion and training strategies on Llama-3.1-8B-Instruct and Qwen2.5-7B-Instruct.}
\label{appendix: ablation_results}
\end{table*}

\section{Application}
\label{appendix: application}

\subsection{Correlation between Resistance and Therapeutic Relationship}
\label{appendix: corr_resistance_and_relationship}

Table~\ref{tab:corr_resistance_wai} shows the Pearson correlation results of both specific resistance subcategory proportions and the overall session resistance level with client-rated working alliance scores across the dimensions of goal agreement, task agreement, emotional bond and overall strength.

\subsection{Proof-of-Concept Experiment}
\label{appendix: poc_experiment}

\paragraph{Power Analysis.} Prior to the experiment, we conducted a sample size estimation using G*Power 3.1 software~\citep{faul2009gpower}. The test family was set to "F-tests," with the statistical test specified as "ANOVA: Repeated measurements, within-between interaction." The effect size was set to 0.25, indicating a medium effect size, with $\alpha$ set to 0.05 and $1-\beta$ to 0.80. The power analysis suggested that a total sample size of 34 participants was needed.

Following the experiment, we performed a post-experiment power analysis using G*Power 3.1 again. With an effect size (Cohen's $f$) of 0.565, $\alpha$ set at 0.05, and a total sample size of 62, the computed statistical power was 1.0, indicating robust power to detect the observed effects.

\paragraph{Participants.}
We recruited 62 counselors from various social media platforms to participate in the experiment. Of these, 28 identified as female, 23 as male, and 11 did not disclose their gender. The average age of participants was 25.80 years (std = 4.87). In terms of counseling experience, 34 had less than 1 year, 10 had 1-2 years, 8 had 2-3 years, 3 had 3-4 years, 3 had 4-5 years, and 4 had more than 5 years of experience. Upon completing the experiment, each participant received a compensation of 80 RMB.

\paragraph{Experiment Design.} 
The study employed a pre-post design. During the pre phase, counselors completed a scenario-based response task independently, without any model-generated feedback. In the post phase, participants in the experimental group responded to the same scenarios with support from model-provided feedback, which included classifications of client utterances by resistance subtype or collaboration type, along with corresponding explanations. The control group completed the same scenarios without receiving any feedback. All counselors engaged with an identical set of 30 scenarios, consisting of two examples for each resistance subtype and four collaborative scenarios, presented in randomized order.

\paragraph{Examples of Pre- and Post-Test Responses.} Table~\ref{tab:example_pre_post_response} showcases examples of pre- and post-test responses, demonstrating that with model-based feedback—including resistance types and detailed explanations—counselors respond to client resistance with enhanced understanding and empathy.

\begin{table*}[t]
    \centering
    \scalebox{0.67}{
    \begin{tabular}{cccc}
    \toprule
    \textbf{Context}   & 
    \textbf{Model Feedback} &
    \textbf{Original Response} & 
    \textbf{Re-written Response}     
    \\ \midrule
    \begin{tabular}[c]{@{}l@{}} T: Perhaps you could try talking to her. \\ C: No, it'll just end up in an argument.\end{tabular} & 
    \begin{tabular}[c]{@{}l@{}} \textit{(Denying - Disagreeing)} \\ The client dismissed the \\ counselor's suggestion, \\ deeming it ineffective.\end{tabular} &
    \begin{tabular}[c]{@{}l@{}} If you stay calm, she might \\ respond more positively. \end{tabular} &
    \begin{tabular}[c]{@{}l@{}}I completely understand your concern. \\ When emotions run high, it can definitely \\ make communication more challenging. \end{tabular} \\ \midrule

    \begin{tabular}[c]{@{}l@{}} T: Sounds like you're holding in your \\ feelings. Doesn't that tire you out? \\ C: I'm used to it. I don't think I need \\ to get along with everyone.\end{tabular} & 
    \begin{tabular}[c]{@{}c@{}} \textit{(Denying - Minimizing)} \\ The client dismissed concerns \\ about emotional suppression \\ and minimized conflict.\end{tabular} &
    \begin{tabular}[c]{@{}c@{}} But over time, these emotions \\ may affect your interactions \\ with colleagues. \end{tabular} &
    \begin{tabular}[c]{@{}c@{}} It's great that you have your own ways \\ of coping. I'm here to support you \\ whenever you're ready to talk more.\end{tabular} \\ \midrule

    \begin{tabular}[c]{@{}l@{}} T: It's common to tell ourselves to try \\ harder but lack the motivation.\\ C: it's common doesn't make it right.\end{tabular} & 
    \begin{tabular}[c]{@{}c@{}} \textit{(Arguing - Challenging)} \\ The client questions the coun-\\selor's view by rejecting the \\ link between "common" and \\ "correct."\\ \end{tabular} &
    \begin{tabular}[c]{@{}c@{}} You could make a plan \\ to overcome difficulties. \end{tabular} &
    \begin{tabular}[c]{@{}c@{}} That's a valid point. Maybe we can \\ take some time to explore the reasons \\ behind it together.\end{tabular} \\ 
    \bottomrule      
\end{tabular}}
\caption{Example re-written responses with our model-based feedback.}
\label{tab:example_pre_post_response}
\end{table*}

\paragraph{Experiments Results}
Table~\ref{tab:corr_resistance_wai} presents the Pearson correlation between session-level resistance behavior proportions and client-rated working alliance scores across goal, task, and bond dimensions.

\begin{table}[b]
\centering
\scalebox{0.65}{
\begin{tabular}{lllll}
\toprule
\textbf{Behaviors} & \textbf{Goal} & \textbf{Task} & \textbf{Bond} & \textbf{Overall} \\
\midrule
Challenging         & -0.1632***    & -0.1717***    & -0.1079**     & -0.1598***     \\
Discounting         & -0.2521***    & -0.2425***    & -0.2312***    & -0.2596***     \\
Blaming             & 0.0337        & 0.0333        & 0.0089        & 0.0277         \\
Disagreeing         & -0.036        & -0.0129       & -0.0481       & -0.0339        \\
Excusing            & 0.0231        & 0.0291        & 0.0046        & 0.0209         \\
Minimizing          & -0.1157**     & -0.0954**     & -0.1032**     & -0.1122**      \\
Pessimistic         & 0.0222        & 0.0228        & -0.0228       & 0.0091         \\
Reluctance          & -0.0054       & 0.0002        & -0.032        & -0.0125        \\
Unwillingness       & 0.0236        & 0.002         & 0.0042        & 0.0107         \\
Minimum Talk        & -0.1192***    & -0.1332***    & -0.1357***    & -0.1385***     \\
Limit Setting       & -0.1662***    & -0.1685***    & -0.1492***    & -0.1733***     \\
Sidetracking        & -0.1035**     & -0.1093**     & -0.0735*      & -0.1032**      \\
Inattention         & -0.0487       & -0.0099       & -0.0542       & -0.0394        \\
\midrule
Resistance          & -0.2222***    & -0.2038***    & -0.2246***    & -0.2321***    \\
\bottomrule
\end{tabular}}
\caption{ Pearson correlations between session-level resistance behavior proportions and client-rated working alliance scores. ***/**/* indicate p-value < .001/.01/.05.}
\label{tab:corr_resistance_wai}
\end{table}

\end{CJK*}
\end{document}